\documentclass[runningheads]{llncs}
\usepackage{graphicx}
\usepackage{cite}
\usepackage{amsmath,amssymb,amsfonts}
\usepackage{algorithmic}
\usepackage{textcomp}
\usepackage{xcolor}

\usepackage{orcidlink}
\renewcommand{\orcidID}[1]{\,\orcidlink{#1}}

\usepackage[capitalise]{cleveref}
\usepackage{caption}
\usepackage{subcaption}
\usepackage{url}
\usepackage{booktabs}
\usepackage{tabularx}
\usepackage{multirow}
\usepackage{multicol}
\usepackage{pseudo}
\usepackage{tikz}
\pseudodefinestyle{fullwidth}{
  begin-tabular = \tabularx{\linewidth}{@{}
    r 
    >{\pseudosetup} 
    X 
    >{\leavevmode\small\color{black!60}} 
    p{,\linewidth} 
    @{}},
  end-tabular = \endtabularx,
  setup-append = \pseudoeq
}
\pseudoset{
  ctfont=\color{black!55}\textit,
  ct-left=//\ ,
  ct-right=,
}
\usepackage{float}
\usepackage{comment}
\usepackage{makecell}

\begin{document}
\title{A Study on Inference Latency for Vision Transformers on Mobile Devices}
%
%
\author{Zhuojin Li\orcidID{0000-0002-8308-0231} \and
Marco Paolieri\orcidID{0000-0001-5110-203X} \and
Leana Golubchik\orcidID{0000-0001-8353-5040}\thanks{This work was supported in part by the NSF CNS-1816887, CCF-1763747, and IIS-1833137 awards.}}
\authorrunning{Li et al.}
%
\institute{University of Southern California, Los Angeles, California, USA \\
\email{\{zhuojinl,paolieri,leana\}@usc.edu}}
\maketitle              
\begin{tikzpicture}[remember picture,overlay]
\node[anchor=north west,shift={(0,0)},minimum width=\paperwidth,minimum height=3.3cm,fill=black!5] at (current page.north west){
\begin{minipage}[t]{0.15\paperwidth}
\vspace{-2mm}%
\raggedleft BibTeX Citation:
\end{minipage}\hspace{5mm}
\begin{minipage}[t]{0.8\paperwidth}\footnotesize
\vspace{-2mm}%
\begin{verbatim}
@inproceedings{LiPG26a,
  author    = {Zhuojin Li and Marco Paolieri and Leana Golubchik},
  title     = {A Study on Inference Latency for Vision Transformers on Mobile Devices},
  booktitle = {Proceedings of {VALUETOOLS} 2024},
  series    = {LNICST}, volume = {663}, pages = {229--251},
  publisher = {Springer}, year = {2026}, doi = {10.1007/978-3-032-06818-7\_13}
}
\end{verbatim}
\end{minipage}
};
\end{tikzpicture}
\begin{abstract}
Given the significant advances in machine learning techniques on mobile devices, particularly in the domain of computer vision, in this work we quantitatively study the performance characteristics of 190 real-world vision transformers (ViTs) on mobile devices.
Through a comparison with 102 real-world convolutional neural networks (CNNs), we provide insights into the factors that influence the latency of ViT architectures on mobile devices.
Based on these insights, we develop a dataset including measured latencies of 1000 synthetic ViTs with representative building blocks and state-of-the-art architectures from two machine learning frameworks and six mobile platforms. Using this dataset, we show that inference latency of new ViTs can be predicted with sufficient accuracy for real-world applications.

\end{abstract}

\section{Introduction}

Recent significant advances in machine learning (ML) techniques on mobile devices, particularly in the domain of computer vision (CV), are enabling real-time on-device processing that was once limited to powerful desktop or cloud servers.
The improvements in mobile hardware and optimized ML frameworks allow the deployment of sophisticated neural architectures to serve complex vision tasks, such as motion analysis and augmented reality.

Moreover, recent progress in Vision Transformers (ViTs) has revolutionized CV, demonstrating outstanding ML accuracy across various CV tasks compared to traditional convolutional neural networks (CNNs)~\cite{dosovitskiy2020image}. 
However, the self-attention mechanism used in ViTs (but not CNNs) is computationally expensive, posing challenges on mobile devices with limited computational and memory resources.
Thus, comprehensively analyzing the inference latency of ViTs on mobile platforms is crucial for leveraging their advanced capabilities while ensuring smooth user experience.
Consequently, we focus on the performance characteristics of ViTs on mobile devices which are insufficiently explored in the existing literature; specifically, we highlight three key factors that distinguish our work from related efforts.

\emph{(1) Inference on Mobile Platforms}: Most existing works evaluate the performance of transformers during training on cloud GPUs~\cite{ivanov2021data,andoorveedu2022tempo,cheng2024thorough} with substantial computational power and scalability; in contrast, mobile platforms have limited memory resources to support complex neural networks and the inference latency is especially critical.
Due to limited support for operations in real-world ViTs on mobile GPUs by current ML frameworks, we focus on mobile CPUs in this paper.
Particularly, heterogeneous CPU cores on a mobile device exhibit distinct performance characteristics; for instance, after quantization, we observe performance improvements on efficient cores but degradation on powerful cores.

\emph{(2) Evolving ViT Architectures}: Related work focusing on mobile platforms is typically restricted to CNNs~\cite{sun2023experimental,baller2021deepedgebench} or NLP Transformers~\cite{panopoulos2023exploring}, while state-of-the-art (SOTA) ViTs have hybrid architectures with distinct performance characteristics.
Previous work~\cite{wang2022towards} studied the performance of 9 ViTs, concluding that ViTs were too expensive for mobile devices; specifically, the fastest ViT was slower than any CNN analyzed. In contrast, we measure latency of 190 real-world ViTs, incorporating novel efficient architecture design; for instance, the smallest ViT evaluated in our work, EfficientViT~\cite{cai2022efficientvit}, achieves lower latency than all the CNNs in \cite{wang2022towards}. Through our study, we provide new insights on how memory formats and activation functions affect inference latency.

\emph{(3) Effects of ML Frameworks}: Existing work \cite{wang2022towards,panopoulos2023exploring} provides performance analysis based on a single ML framework, i.e., TensorFlow Lite (TFLite).
However, since most SOTA ViTs are implemented in PyTorch, we mainly study the performance characteristics of ViTs on PyTorch Mobile, and also conduct comparisons with TFLite.
We note that distinct implementations across ML frameworks can substantially impact inference latency; e.g., a convolution operation can exhibit significantly different latency on the same mobile device when implemented by different ML frameworks.

In this paper, we characterize the factors that influence the latency of ViTs on mobile platforms and develop a dataset including measured latencies of synthetic ViTs with representative building blocks and SOTA architectures. Using this dataset, we show that inference latency of ViTs can be predicted with sufficient accuracy for many applications of interest, including:

(i) \emph{Neural Architecture Search (NAS)}~\cite{zoph2016neural}, which automates the design of neural architectures to obtain good tradeoffs between accuracy and efficiency; latency prediction can save the cost of deployment to mobile devices and ensure that the selected neural architecture satisfies the latency constraints on a target device.
In line with the principle of NAS, we construct latency predictors trained on synthetic ViTs and demonstrate their high accuracy on 100 candidate ViT architectures sampled from the same search space.

(ii) \emph{Collaborative} (or \emph{Split}) \emph{Inference}~\cite{kang2017neurosurgeon}, which offloads partial model computation to cloud servers with powerful computational resources while preserving privacy; latency prediction for each part of the model facilitates determining optimal model partition with tradeoff between the saved local computation time and additional transmission cost.
Accordingly, we evaluate our pre-trained models on 190 real-world ViTs, illustrating that our predictors can accurately estimate latency of novel ViT architectures.

Our main contributions are summarized as follows:
\begin{itemize}
    \item We quantitatively compare the performance characteristics of 190 real-world ViTs and 102 CNNs on mobile platforms (\cref{sec:comparison}), illustrating their differences in latency, performance bottlenecks, and memory consumption. Based on this comparison of performance characteristics, we develop insights into the fundamental causes of latency patterns observed in ViTs (\cref{sec:insight}), including the effects of memory formats, selection of activation functions, and implementations provided by ML frameworks.

    \item Based on this thorough understanding of latency characteristics, we design a search space to generate synthetic ViTs with representative building blocks used in SOTA efficient ViTs (\cref{sec:synthetic_design}). We release the resulting ViT latency dataset~\cite{dataset}, containing profiling information for 1000 synthetic ViTs and 190 real-world ViTs on 6 mobile platforms and 2 mainstream ML frameworks, across different CPU core combinations and data representations, which can be used by researchers and developers for performance analysis.

    \item To demonstrate the applicability of our dataset, we train ML latency predictors from measurements of 900 synthetic ViTs on 6 mobile devices. From systematic evaluation of these predictors on both 100 synthetic (\cref{sec:result_synthetic}) and 190 real-world ViTs (\cref{sec:result_realworld}), we show that simple ML predictors trained on our dataset are sufficiently accurate for practical applications such as NAS (i.e., for synthetic ViTs, errors of 4.4\% on PyTorch Mobile and 4.8\% on TFLite for mobile CPUs, and errors of 2.1\% on PyTorch Mobile and 8.9\% on TFLite for mobile GPUs) and collaborative inference (i.e., errors of 8.2\% on PyTorch Mobile and 6.1\% on TFLite for real-world ViTs on mobile CPUs).
\end{itemize}

\section{Background and Experimental Setup}\label{sec:background}

\subsection{Vision Transformers}\label{sec:background_vit}

Vision Transformers (ViTs) are a class of neural architectures intended for vision tasks that leverage a \emph{self-attention} mechanism, originally introduced by transformers for NLP tasks~\cite{vaswani2017attention}.
Initially proposed in \cite{dosovitskiy2020image}, ViTs have fundamentally changed the domain of computer vision, emerging as an alternative to more traditional CNNs.

\begin{figure}[t]
    \centering
	\includegraphics[width=\linewidth]{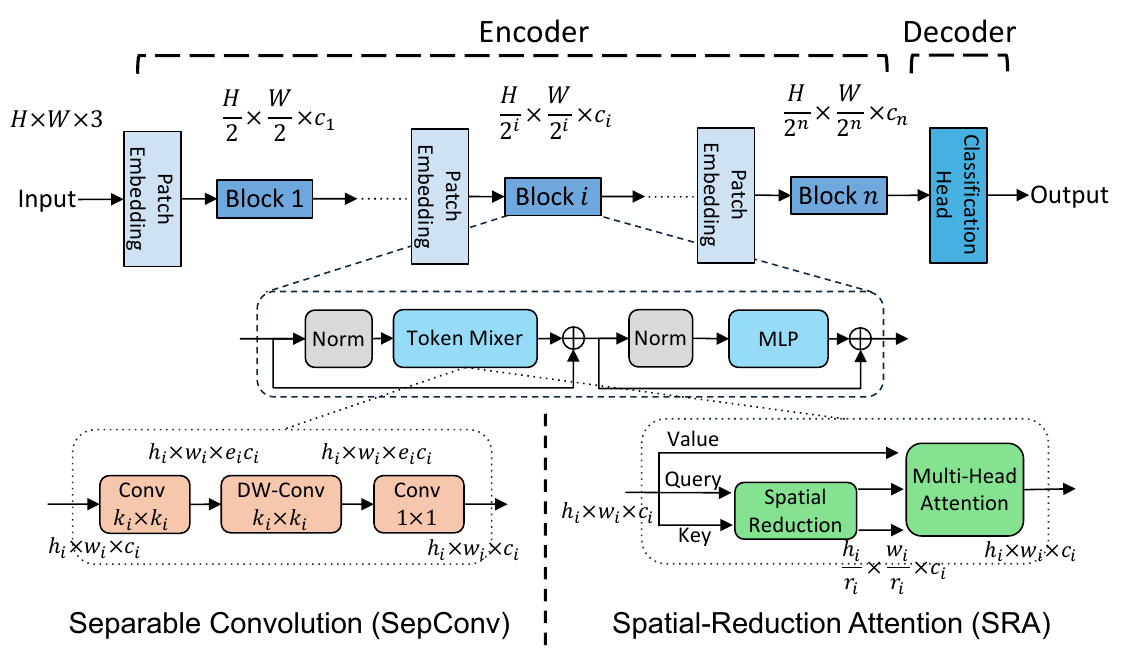}\vspace{-.5em}
	\caption{Search Space Design for Synthetic ViTs}\label{fig:search_space_design}
    \vspace{-1em}
\end{figure}

At a high level, the architecture of ViTs consists of an encoder and a decoder, as demonstrated in \cref{fig:search_space_design} for our synthetic ViT design (\cref{sec:result_synthetic}).
The encoder represents the input image as a set of embedding vectors and performs a series of transformations; the results are passed to the decoder, which generates the final output.
Specifically, the encoder divides an input image into patches (e.g., 16x16 pixels~\cite{dosovitskiy2020image}); each patch is then flattened into a 1D vector and projected to a high-dimensional embedding space through linear transformation, as illustrated by the patch embedding block in \cref{fig:search_space_design}.
Similarly to word embeddings (also known as tokens) in NLP transformers, the embedding vectors for image patches are passed through multiple \emph{multi-head self-attention}~\cite{vaswani2017attention} blocks, also known as a \emph{token mixers} (see \cref{fig:search_space_design}).
The general idea of the self-attention mechanism is to allow each embedding vector (e.g., an image patch) to exchange information with other embeddings; specifically, each embedding vector ``pays attention" to others in the input sequence based on their ``relevance".
This is a key difference between ViTs and CNNs: while ViTs consider global relationships between pairs of embeddings, CNNs hierarchically process local spatial details (e.g., edge) before aggregating into high-level features (e.g., shapes).

The multi-head self-attention block (1) performs linear transformations on input embeddings to obtain query, key, and value matrices ($Q, K, V$) with embedding length $c_i$, (2) computes attention scores $QK^{\top}$ (i.e., the relevance of each embedding to others) and (3) outputs a weighted sum of values based on attention weights:  $\text{softmax}(\frac{QK^{\top}}{\sqrt{c_i}})V$.
The latency of this block is primarily dominated by the computation of linear transformations and matrix multiplications.
Specifically, for an input sequence of $N$ tokens, each represented as an embedding vector of length $D$, the computational complexity of the linear transformations to obtain $N \times D$ query, key, and value matrices ($Q, K, V$) is $\mathcal{O}(ND^2)$.
Additionally, in the multi-head attention layer, the computational complexity of matrix multiplication, $QK^{\top}$, used to obtain $N \times N$ attention scores is $\mathcal{O}(N^2 D)$. This quadratic cost is due to  the need to compute scores for each pair of tokens in the input sequence. Also, the matrix multiplication between the normalized attention weights, $\text{softmax}(\frac{QK^{\top}}{\sqrt{c_i}})$, and the value matrix, $V$, to produce an $N \times D$ output embedding matrix has a computational complexity of $\mathcal{O}(N^2 D)$.
Consequently, as the number of tokens, $N$, increases, the complexity of an attention block grows \textit{quadratically}, which significantly impacts latency.

Notably, recent efficient ViT designs~\cite{yu2023metaformer} tend to use convolution blocks alternatively in the early layers for feature extraction, such as \emph{Separable Convolution} (SepConv)~\cite{howard2017mobilenets} that significantly reduces the computational cost by decomposing a standard convolution operation into \emph{depthwise convolution} (DWConv, where each filter is applied to a separate input channel rather than all input channels in a standard convolution) and \emph{pointwise convolution} (convolution with $1 \times 1$ filter).
In addition, similarly to the hierarchical structures in CNNs, recent real-world ViTs~\cite{wang2021pyramid,liu2021swin} commonly process features at multiple scales across different blocks. Specifically, the patch embedding~\cite{dosovitskiy2020image} divides the input feature map of shape $h \times w$ into patches of size $2 \times 2$, yielding $\frac{h}{2} \times \frac{w}{2}$ embedding vectors, with increasing embedding lengths in deeper blocks.
The decoder (or neck) of a ViT can be configured for a specific task, such as classification head for image classification or detection head for object detection.

\subsection{Deployment of ViTs on Mobile Platforms}\label{sec:background_mobile_setup}

Compared to cloud servers, mobile devices are equipped with limited RAM (e.g., 6 GB on iPhone 15 instead of 80 GB GPU memory on Nvidia H100). Thus, deployment of complex ViTs becomes challenging due to their substantial memory requirement during inference. For instance, the object detection ViT model DETR-ResNet101~\cite{carion2020end} consumes 5.0 GB mobile memory for a single image of $512 \times 1333$ pixels, and higher input resolutions require more memory (\cref{sec:comparison_memory}).
Additionally, memory access plays an important role in inference latency (\cref{sec:comparison_latency}), which is crucial for mobile platforms with demanding user experience but constrained resources.

To systematically study the critical factors of inference latency, we collect 190 real-world ViTs from \textit{Timm}~\cite{rw2019timm} and \textit{HuggingFace's Transformers}~\cite{wolf-etal-2020-transformers}, and convert these models into PyTorch Mobile format to facilitate deployment across 6 mobile platforms summarized in \cref{table:mobile_platforms}.
Specifically, we deploy all the ViTs on mobile CPUs because many common operations in ViTs are currently unavailable in ML frameworks for mobile GPUs, e.g., the \textit{roll} operation in Swin~\cite{liu2021swin}.
Following \cite{li2024benchmark}, for each ViT, we measure the latencies of a randomly initialized image across various combinations of CPU cores, ensuring that each core has a thread assigned to it.
Additionally, we evaluate the effects of \emph{quantization}, a popular optimization approach to reducing computational cost and memory consumption; we note that 64 of the 190 real-world ViTs cannot be quantized due to operations unsupported by PyTorch Mobile.
To investigate the impact of ML frameworks on inference latency, we also deploy TensorFlow models in TensorFlow Lite (TFLite); we note that only 25 of the 190 ViTs have TensorFlow implementations available.

\begin{table}[t]\centering \footnotesize
\setlength{\tabcolsep}{0.5em}
\renewcommand\arraystretch{1.1}
\centering
\begin{tabular}{c c c}\toprule
\textbf{Platform (SoC)} & \textbf{CPU Cores} & \textbf{GPU} \\
\toprule
\makecell{Google Pixel 4 \\ (Snapdragon 855) } & \makecell{1x Large (2.84 GHz),\\ 3x Medium (2.32 GHz),\\ 4x Small (1.80 GHz)} & Adreno 640 \tabularnewline
\cmidrule(lr){1-3}
\makecell{Motorola One Fusion \\ (Snapdragon 710)} & \makecell{2x Large (2.20 GHz),\\ 6x Small (1.70 GHz)} & Adreno 616 \tabularnewline
\cmidrule(lr){1-3}
\makecell{Samsung Galaxy S10 \\ (Exynos 9820) } & \makecell{2x Large (2.73 GHz),\\ 2x Medium (2.31 GHz),\\ 4x Small (1.95 GHz)} & Mali G76 \tabularnewline
\cmidrule(lr){1-3}
\makecell{Samsung Galaxy A03s \\ (Helio P35)} & \makecell{4x Large (2.30 GHz),\\ 4x Small (1.80 GHz)} & PowerVR GE8320 \tabularnewline
\cmidrule(lr){1-3}
\makecell{Apple iPhone 12 \\ (A14 Bionic)} & \makecell{2x Large (2.31 GHz),\\ 4x Small (1.80 GHz)} & Apple-designed (4 core) \tabularnewline
\cmidrule(lr){1-3}
\makecell{Apple iPhone XS \\ (A12 Bionic)} & \makecell{2x Large (2.49 GHz),\\ 4x Small (1.52 GHz)} & Apple-designed G11P \tabularnewline

\bottomrule
\end{tabular}
\vspace{1mm}
\caption{Mobile Platforms in Our Study}
\label{table:mobile_platforms}\vspace{-.75em}
\end{table}

\section{Evaluation on Real-World ViTs}

In this section, we study the performance characteristics of 190 real-world ViTs from our dataset~\cite{dataset}. We begin by comparing with real-world CNNs (\cref{sec:comparison}) and provide insights into inference latency based on our findings (\cref{sec:insight}).

\subsection{Comparison between CNNs and ViTs}\label{sec:comparison}

\begin{figure}[t]
    \centering
	\includegraphics[width=.62\linewidth]{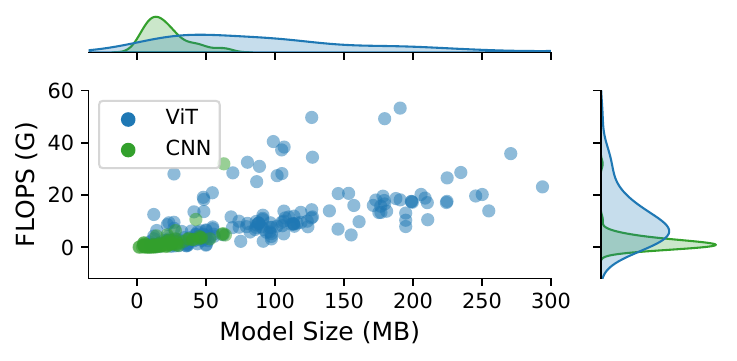}\vspace{-.5em}
	\caption{Overview of Evaluated Architectures}\label{fig:comparison_flops_size}
    \vspace{-.75em}
\end{figure}

Firstly, we analyze performance characteristics of real-world ViTs on PyTorch Mobile through a comparison with 102 real-world CNNs studied in \cite{li2024benchmark}.

\subsubsection{Latency}\label{sec:comparison_latency}

\begin{figure}[t]
    \centering
	\includegraphics[width=.62\linewidth]{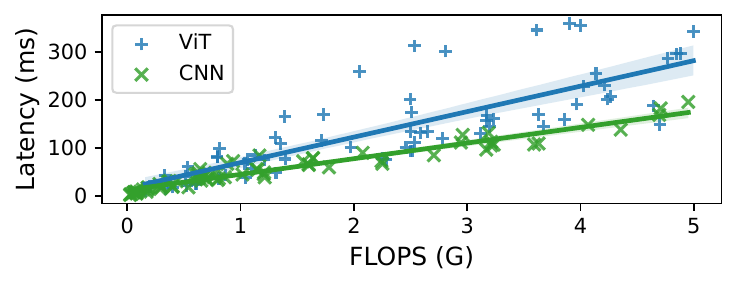}\vspace{-.5em}
	\caption{End-to-End Latency Comparison}\label{fig:comparison_e2e_latency}
    \vspace{-.75em}
\end{figure}

\begin{figure}[t]
    \centering
	\includegraphics[width=.8\linewidth]{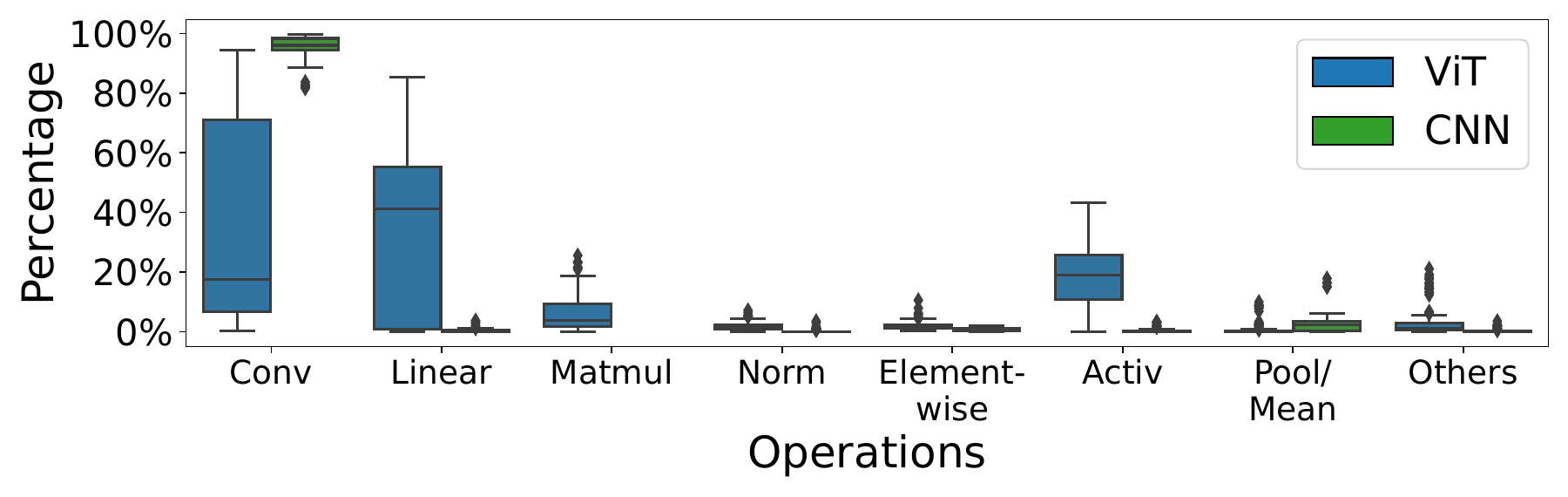}\vspace{-.5em}
	\caption{Latency Breakdown Comparison}\label{fig:comparison_ops_latency}
    \vspace{-.75em}
\end{figure}

\begin{figure}[t]
    \centering
	\includegraphics[width=.62\linewidth]{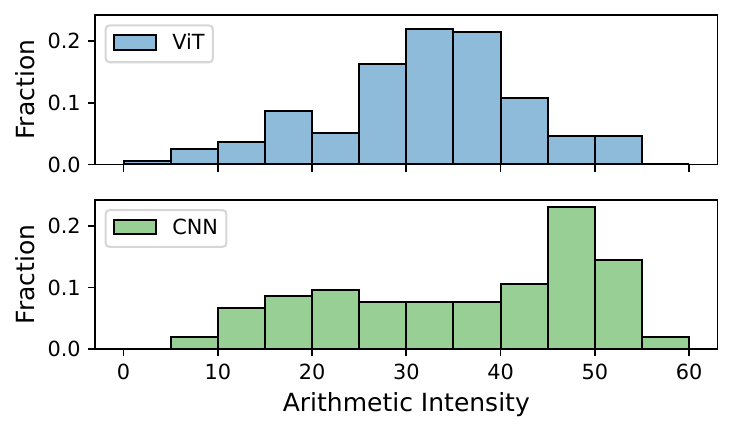}\vspace{-.5em}
	\caption{Histograms of Arithmetic Intensity}\label{fig:comparison_intensity}
    \vspace{-.75em}
\end{figure}

\cref{fig:comparison_flops_size} illustrates the distribution of the number of floating-point operations (FLOPs) and model sizes for common ViTs and CNNs.
Generally, ViTs exhibit higher FLOPs than CNNs, primarily due to fundamental architectural differences: specifically, ViTs apply self-attention mechanisms that compute the relevance between all pairs of patches in the input image, leading to a complexity of $\mathcal{O}(H^2W^2)$ for an image with dimensions $H \times W$ (\cref{sec:background_vit}); in contrast, CNNs rely on convolution layers that apply fixed-size filters to each local region of the input, resulting in a lower complexity of $\mathcal{O}(HW)$.
%
We measure the inference latency for CNNs and ViTs with comparable FLOPs (i.e., less than 5~GFLOPs); as presented in \cref{fig:comparison_e2e_latency}, where we fit linear regression models to indicate the trend between FLOPs (estimated by \emph{ptflops}~\cite{ptflops}) and latency, \emph{ViTs consistently exhibit longer inference times than CNNs with similar FLOPs}; e.g., the ViT with 5.00 GFLOPs incurs 1.75x latency of the CNN with 4.95 GFLOPs.

To better understand inference latency performance, we profile the latency of each type of operation and depict their contribution to overall latency in \cref{fig:comparison_ops_latency}. 
We observe two main differences between ViTs and CNNs: 
(1) While the majority of end-to-end latency in CNNs is due to convolution operations, a significant portion of the end-to-end latency in ViTs is due to linear operations, as these are the crucial components of self-attention blocks.
%
(2) A larger proportion of the end-to-end latency in ViTs is attributed to activation operations, primarily Gaussian Error Linear Unit (GELU) activations~\cite{hendrycks2016gaussian}.
We observe that the computation of GELU activation is \emph{approximated} differently depending on its input values, significantly affecting latency (as detailed in \cref{sec:insight_activation}).
The variability in approximations makes the FLOPs estimation (which is calculated independently of the input values) inaccurate to represent actual latency of ViTs.
%
%

\emph{Takeaway:} ViTs generally exhibit higher latency than CNNs with similar FLOPs, suggesting that ViTs may be less suitable for real-time applications on mobile platforms where low latency is critical.
Existing literature has demonstrated that FLOPs is not an accurate latency proxy for CNNs~\cite{ma2018shufflenet}, and it is even less reliable for ViTs due to the variable latency of GELU activations for different input values.

\subsubsection{Performance Bottleneck}\label{sec:comparison_bound}

Existing literature~\cite{ivanov2021data,kao2023flat} suggests that self-attention blocks are typically less memory-efficient compared to convolutions.
To confirm this on mobile platforms, \cref{fig:comparison_intensity} compares their distributions of \emph{arithmetic intensity}~\cite{boroumand2021google} on Pixel 4, which is defined as the ratio of total number of floating-point instructions divided by the amount of memory traffic (both reported by Performance Monitoring Unit (PMU) counters on ARM processors).
We observe that arithmetic intensity of ViTs is generally lower than that of CNNs, indicating a greater tendency for \emph{memory-bound} performance on mobile devices.

\begin{figure}[t]
    \centering
 	\begin{subfigure}[t]{.5\linewidth}
        \centering
    	\includegraphics[width=\linewidth]{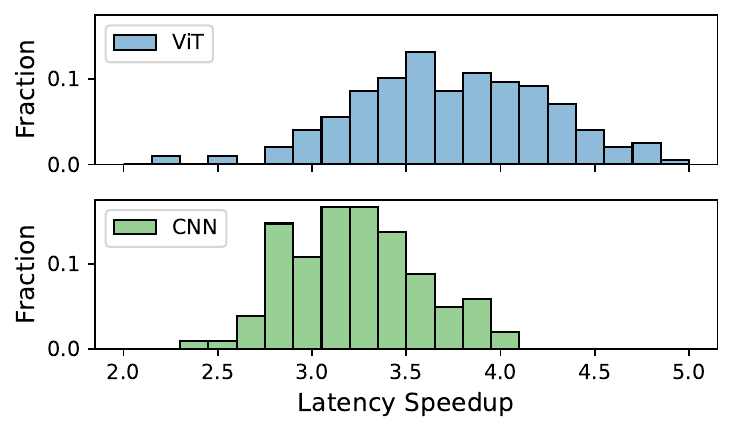}\vspace{-.5em}
    	\caption{Memory Bandwidth Increase}\label{fig:speedup_bandwidth}
	\end{subfigure}
    \hspace{-.8em}
    \begin{subfigure}[t]{.5\linewidth}
        \centering
    	\includegraphics[width=\linewidth]{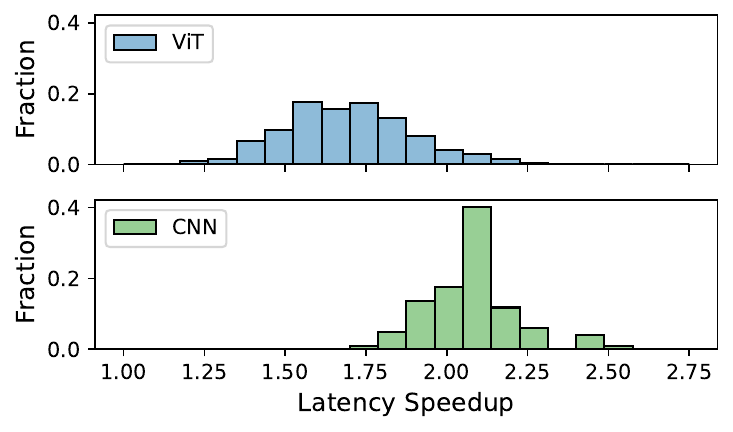}\vspace{-.5em}
    	\caption{CPU Clock Speed Increase}\label{fig:speedup_frequency}
	\end{subfigure}
	\caption{Speedup Histograms Comparing Latency between (a) Highest and Lowest Memory Bandwidth and (b) Fastest and Slowest CPU Clock Speed}\label{fig:speedup_bandwidth_and_frequency}
    \vspace{-.75em}
\end{figure}

To confirm this memory bottleneck, we conduct additional experiments by varying (1) memory frequencies (including L3 cache, last-level cache controller, and DDR RAM) controlled by Dynamic Voltage and Frequency Scaling (DVFS) governors~\cite{qualcomm_linux_performance_guide} or (2) CPU frequencies when running inference with 1 large and 3 medium cores on Pixel 4.
As depicted in \cref{fig:speedup_bandwidth}, increasing memory frequencies from the lowest to the highest results in a more substantial acceleration for ViTs as compared to CNNs; for example, 75\% of ViTs achieve speedups over 3.40x, compared to only 28\% of CNNs.
On the other hand, as illustrated in \cref{fig:speedup_frequency}, boosting CPU clock from the slowest (0.83 GHz) to the fastest (2.84 GHz) speed provides less reduction in latency for ViTs as compared to CNNs; for example, 89\% of ViTs show speedups less than 1.93x, compared to only 12\% of CNNs.

\emph{Takeaway:} Both experiments suggest that ViTs are more likely to be memory-bound than CNNs and therefore require higher memory bandwidth on mobile platforms due to larger memory access demands.

\subsubsection{Memory Consumption}\label{sec:comparison_memory}

\begin{figure}[t]
    \centering
	\includegraphics[width=.62\linewidth]{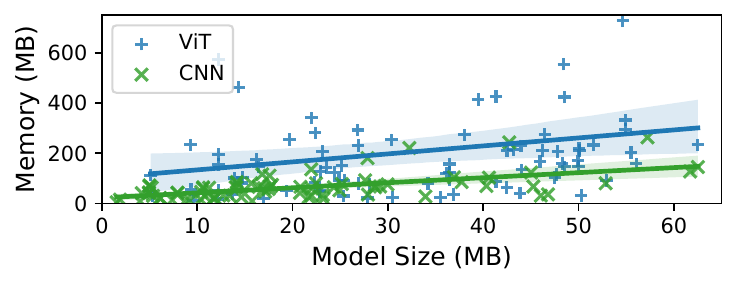}\vspace{-.5em}
	\caption{End-to-End Memory Requirement Comparison}\label{fig:comparison_size_mem}
    \vspace{-.75em}
\end{figure}

\begin{figure}[t]
    \centering
	\includegraphics[width=.75\linewidth]{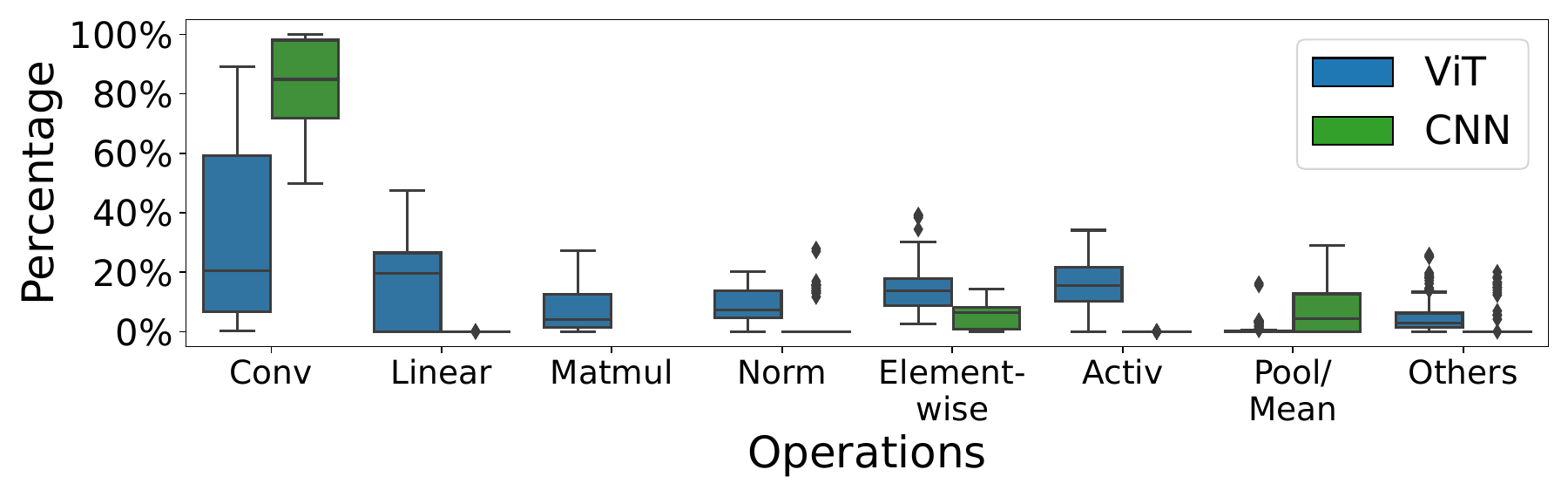}\vspace{-.5em}
	\caption{Memory Breakdown Comparison}\label{fig:comparison_ops_mem}
    \vspace{-1em}
\end{figure}

\cref{fig:comparison_flops_size} shows that ViTs typically exhibit larger model sizes than CNNs, because the self-attention blocks consist of more complex parameters than convolution layers.
We measure the memory consumption of intermediate tensors during inference reported in PyTorch Mobile for ViTs and CNNs with comparable model sizes (i.e., less than 65 MB); as depicted in \cref{fig:comparison_size_mem}, the memory consumption is greater for ViTs.

Particularly, we also note different contributions to memory consumption across various types of operations, as depicted in \cref{fig:comparison_ops_mem}.
In CNNs, element-wise and pooling operations contribute to a minor portion of the end-to-end latency (i.e., less than 1.9\% on average) but to a higher portion of memory consumption (e.g., 6.6\% for pooling and 3.9\% for add).
In ViTs, linear operations account for a smaller portion of memory use than latency; normalization, element-wise and activation operations contribute to a significant amount of memory consumption (e.g., 9.4\% for GELU, 7.6\% for add, 5.3\% for layernorm, and 4.9\% for softmax). These operations are fundamental components of self-attention blocks.

In addition, we observe that memory consumption scales differently for ViTs than CNNs with the increase of input resolution. For example, we study two typical architectures of ViTs and CNNs: Vanilla ViT-S (with patch size 32)~\cite{dosovitskiy2020image} and ResNet-18 (with width scale 0.5)~\cite{he2016deep}. The memory consumption of Vanilla ViT-S is 14\% higher than ResNet-18 at an input resolution of 224x224 pixels, but the difference grows to 43\% at 512x512 pixels.

\emph{Takeaway:} The memory requirements of ViTs generally exceed those of CNNs with comparable model sizes, and this memory consumption scales more rapidly as ViT input resolution increases. Thus, careful consideration is required when choosing ViTs over CNNs in memory-constrained environments.

\subsection{Understanding ViT Inference Latency: Insights}\label{sec:insight}

Next, we elaborate on insights from our study of inference latency of ViTs, which are taken into consideration in designing our synthetic ViTs (\cref{sec:synthetic_design}).

\subsubsection{Memory Format}\label{sec:insight_memformat}

\begin{figure}[t]
    \centering
	\includegraphics[width=.62\linewidth]{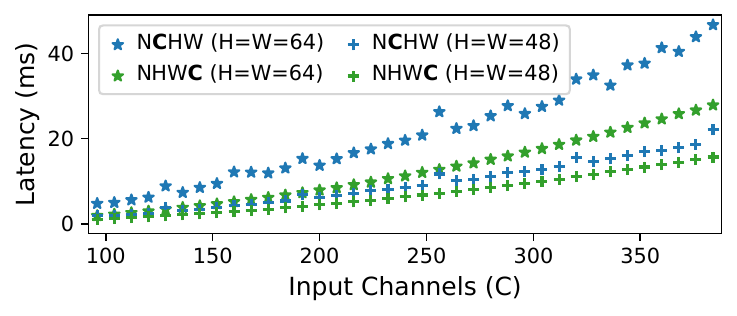}\vspace{-.5em}
	\caption{Effects of Memory Format}\label{fig:memformat_performance}
    \vspace{-.75em}
\end{figure}

%
In profiling latency of ViTs on PyTorch Mobile, we observe a significant number of \emph{transpose} and \emph{reshape} operations; this is attributed to the difference in input shapes between convolution operations (requiring 4-dimensional tensors) and self-attention blocks (expecting 3-dimensional tensors).
Specifically, PyTorch utilizes \emph{channel-first} memory format for 4-dimensional tensors (i.e., $[N, C, H, W]$ for batch size, number of channels, height and width in an image). However, before forwarding to self-attention blocks, the image is divided into multiple patches (i.e., small pixel groups) requiring (1) transpose and reshape of the tensor into \emph{channel-last format}, as $[N, H \times W, C]$ and (2) division into image patches of $P^2$ pixels, as $[N, \frac{H \times W}{P^2}, P^2 \times C]$.
Although the transpose and reshape operations contribute to an insignificant portion of the end-to-end latency, they can affect the execution time of a subsequent convolution operation due to changing the memory format of the input tensor.

\cref{fig:memformat_performance} presents the effects of memory format on the latency of convolution operations in PyTorch Mobile.
As can be seen, memory format can significantly affect latency, leading to average speedups of 2.21x and 1.58x for input tensors in channel-last format (NHWC, in blue) compared to channel-first format (NCHW, in green) for tensor shapes of $64 \times 64$ and $48 \times 48$, respectively. This distinction is due to the computing library XNNPACK~\cite{xnnpack} providing different implementations of convolution based on the input and output memory formats.

\emph{Takeaway:} Optimizing the memory format is crucial in the implementation of ViTs, as leveraging an efficient memory format can significant improve the latency of convolution layers. This underscores the importance of sampling synthetic ViTs with different memory formats in our dataset (\cref{sec:synthetic_design}).

\subsubsection{Activation Functions}\label{sec:insight_activation}

\begin{figure}[t]
    \centering
	\includegraphics[width=.62\linewidth]{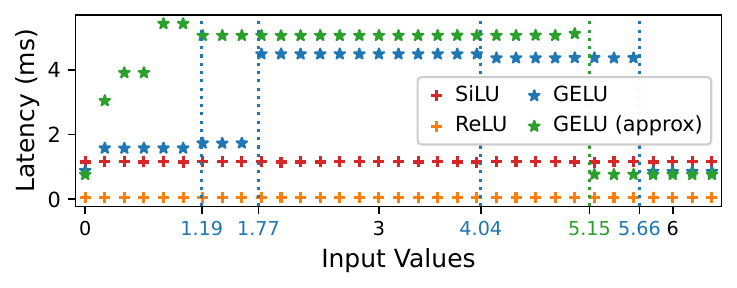}\vspace{-.5em}
	\caption{Latency of Activation Functions}\label{fig:activ_performance}
    \vspace{-.75em}
\end{figure}

In \cref{sec:comparison_latency}, we identify GELU activation operations as significant contributors to latency; here we further study the effects of input values on GELU's latency.
\cref{fig:activ_performance} depicts latency measurements for common activation operations used in vision models, based on a $56 \times 56$ input tensor with variable input channels (indicated on the $x$ axis); here, a clear discontinuity can be observed for GELU (blue crosses), compared to SiLU~\cite{elfwing2018sigmoid} (green dots) and ReLU (red dots).
The GELU operation is defined in~\cite{hendrycks2016gaussian} as:
$\text{GELU}(x) = \frac{x}{2} [1 + \text{erf}(x / \sqrt{2})]$, where the Gaussian error function \emph{erf} is computed by the standard C library \emph{libm} (for basic mathematical functions).
By inspecting the source code, we find that \emph{erf} is computed differently depending on its (absolute) input values;
the discontinuities observed in the figure correspond to the following GELU input values \{1.19, 1.77, 4.04, 5.66\} found in the code.
We observe that this specific approximation leads to substantial differences in GELU's latency; for instance, the latency for an input value of 2 is 2.85x longer than that of an input value of~1.
PyTorch also provides an approximate implementation of GELU based on \emph{tanh}: $\text{GELU}(x) \approx \frac{x}{2} (1 + \text{tanh}(\sqrt{2/\pi} (x + 0.044715 x^3)))$. However, as depicted in \cref{fig:activ_performance}, the latency of approximate GELU (green crosses) also depends on input values, similarly due to the computation of \emph{tanh} in \emph{libm}.
%

\emph{Takeaway:} Latency of the activation function in ViTs cannot be captured by FLOPs; since the values in the input image and model weights influence the inputs to the GELU activation functions, they affect inference latency of ViTs.

\subsubsection{Effects of ML Frameworks}\label{sec:insight_frameworks}

\begin{figure}[t]
    \centering
 	\begin{subfigure}[b]{.5\linewidth}
        \centering
    	\includegraphics[width=\linewidth]{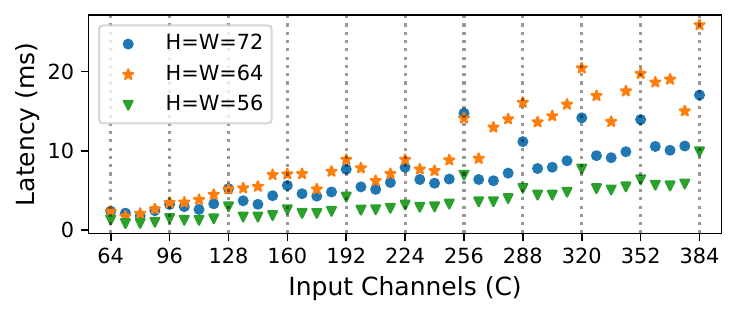}\vspace{-.5em}
    	\caption{PyTorch Mobile}\label{fig:dwconv_performance_torch}
	\end{subfigure}
     \hspace{-.8em}
 	\begin{subfigure}[b]{.5\linewidth}
        \centering
    	\includegraphics[width=\linewidth]{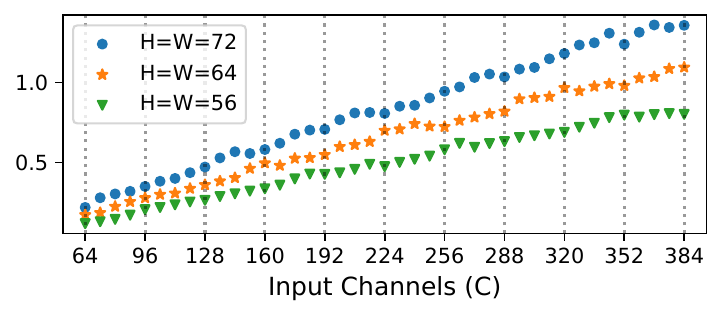}\vspace{-.5em}
    	\caption{TFLite}\label{fig:dwconv_performance_tflite}
	\end{subfigure}
	\caption{Latency of DWConv in Different ML Frameworks}\label{fig:dwconv_performance_comparison}
    \vspace{-.75em}
\end{figure}

\cref{fig:dwconv_performance_torch} illustrates latency measurements for depthwise convolution (DWConv) operations (with various input channels) in PyTorch Mobile. Here, we observe a \emph{non-linear} increase in latency, with spikes occurring when the number of input channels is a multiple of 32.
Notably, the FLOPs of a DWConv operation with filter size $k \times k$ is $H \times W \times C \times k^2$, which grows linearly with number of input channels $C$ and thus does not serve as an accurate proxy metric for inference latency~\cite{li2024benchmark}.
Furthermore, the latency of input shapes $64 \times 64$ is surprisingly greater than that of input shapes $72 \times 72$ for the same number of input channels.
By analyzing the source code of the computing library XNNPACK, we observe a significant cost associated with converting memory formats before and after the convolution computation, due to the different formats between PyTorch Mobile and XNNPACK.
%
%
As a comparison, in \cref{fig:dwconv_performance_tflite} DWConv in TFLite is significantly faster and exhibits a linear increase in performance with the number of input channels, because TFLite and its computing library have the same memory format (i.e., no need to convert).

\begin{figure}[t]
    \centering
 	\begin{subfigure}[t]{.5\linewidth}
        \centering
    	\includegraphics[width=\linewidth]{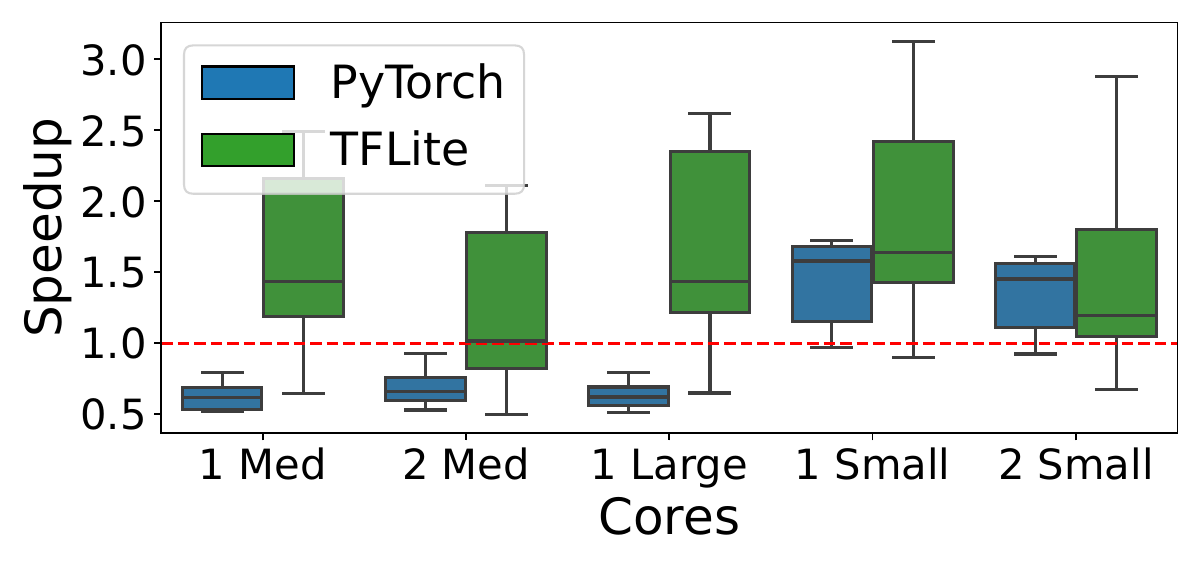}\vspace{-.5em}
    	\caption{End-to-End Latency}\label{fig:quant_framework_e2e}
	\end{subfigure}
    \hspace{-.8em}
    \begin{subfigure}[t]{.5\linewidth}
        \centering
    	\includegraphics[width=\linewidth]{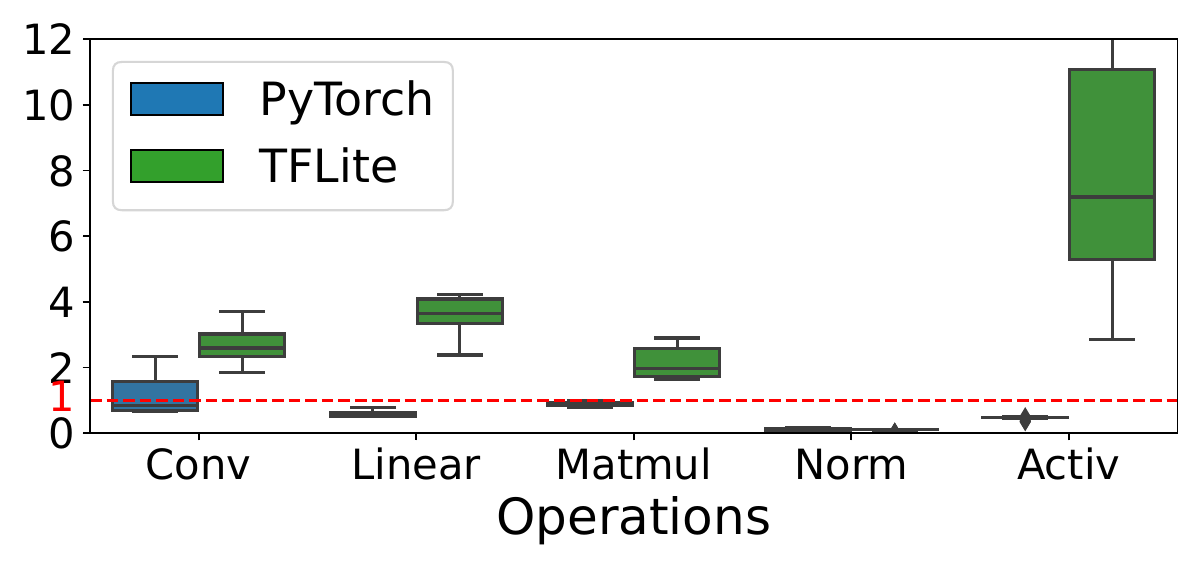}\vspace{-.5em}
    	\caption{Operation (1 Large Core)}\label{fig:quant_framework_ops}
	\end{subfigure}
	\caption{Effects of Quantization (25 Real-World ViTs)}\label{fig:quantization_framework}
    \vspace{-1em}
\end{figure}

In addition, we study the effects of \emph{quantization} in different ML frameworks, a common technique for reducing latency and memory usage through lower precision of data representations.
\cref{fig:quantization_framework} compares the latency speedup on Pixel 4 achieved by quantization for 25 ViTs implemented in both PyTorch Mobile and TFLite.
As can be seen in \cref{fig:quant_framework_e2e}, after quantization in PyTorch Mobile, ViTs achieve acceleration on small cores but not on large or medium cores.
To understand this behavior, \cref{fig:quant_framework_ops} depicts the speedup for each type of operation on a large core.
Notably, there is a performance degradation of linear operations in PyTorch Mobile, due to the use of a different computing library QNNPACK~\cite{qnnpack} for integer data representations.
We observe that the library for floating-point computation (XNNPACK) offers more efficient implementations for linear operations based on Neon SIMD instructions, outperforming quantized implementations in QNNPACK on powerful (large and medium) cores.
Another difference between ML frameworks is the activation function, which shows substantial performance improvement in TFLite, but exhibits performance degradation in PyTorch Mobile.
In addition, we observe a significant performance drop for normalization operations in both frameworks, due to the overhead of input scaling and lack of efficient quantized implementations~\cite{li2023predicting}.

\emph{Takeaway:} Various computing libraries within ML frameworks can lead to significant variations in inference latency; this motivates the inclusion of latency measurements from multiple ML frameworks in our dataset (\cref{sec:synthetic_design}).

\section{Construction of a ViT Latency Prediction Dataset}\label{sec:synthetic}

In this section, we develop our methodology for constructing a search space for sampling synthetic ViTs with representative building blocks (\cref{sec:synthetic_design}) and showcase latency measurements for 1000 synthetic ViTs (\cref{sec:synthetic_measurement}).
We also demonstrate the applicability of our dataset~\cite{dataset} by showing that simple ML predictors (\cref{sec:methodology_predictor}) trained on synthetic ViTs can lead to accurate latency predictions on both synthetic (\cref{sec:result_synthetic}) and real-world (\cref{sec:result_realworld}) ViTs.

\subsection{Search Space Design}\label{sec:synthetic_design}

We sample synthetic ViTs with the hierarchical architecture depicted in \cref{fig:search_space_design}, for input image height $H$ and width $W$ chosen from the set $\{224, 256\}$, as commonly used in vision datasets; we incorporate six blocks ($n=6$) in the architecture so that the encoder outputs a tensor shape of $\frac{H}{64} \times \frac{W}{64}$ for the classification head. The embedding lengths $\{c_i\}$ of these blocks are in an increasing pattern, enabling more detailed feature representations in the deeper blocks:
$
c_1 \in [16, 32],
c_2 \in [32, 80],
c_3 \in [64, 192],
c_4 \in [192, 384],
c_5 \in [256, 768],
c_6 \in [384, 1024]
$.
In order to reduce the computational cost and sample more configurations, we merge the first $k \in [2, 4]$ blocks by removing the first $k - 1$ blocks and assigning patch size of $2^k \times 2^k$ for the $k$-th block, such that the input resolutions over the rest of the blocks remain the same.
%
%
Following the trend of hybrid building blocks~\cite{yu2023metaformer}, we select either \emph{SepConv} or \emph{Attention} as the token mixer (as illustrated in \cref{sec:background_vit}); this design also allows the synthetic ViTs to incorporate tensors with different memory formats (\cref{sec:insight_memformat}). 
%
%
The choices of parameters in each block include: (1) \emph{batchnorm} or \emph{layernorm} as the normalization, both commonly applied in the literature;
(2) GELU or SiLU as the activation function, exhibiting distinct performance characteristics (\cref{sec:insight_activation});
(3) the MLP expansion ratios in the six blocks are sampled from ranges $[1, 4], [2, 10], [2, 10], [1, 4], [1, 4], [1, 2]$ in that order, allowing various dimensions of the intermediate tensors within the attention block.
%
In addition, for SepConv blocks, the convolution filter shape is selected from sets $\{1, 3, 5, 7\}$; the number of input channels to DWConv is selected using expansion ratios of up to $e_i = 8, 8, 8, 8, 4, 2$, respectively.
%
For Attention blocks, we use up to 12 attention heads; we use Spatial-Reduction Attention (SRA)~\cite{wang2021pyramid} in the first three blocks to save computational cost by reducing the input scale to multi-head attention, with sequence reduction ratios sampled from ranges $[2, 16], [1, 4], [1, 2]$.

\subsection{Latency Breakdown for CPU and GPU Execution}\label{sec:synthetic_measurement}

\begin{figure}[t]
    \centering
	\includegraphics[width=.75\linewidth]{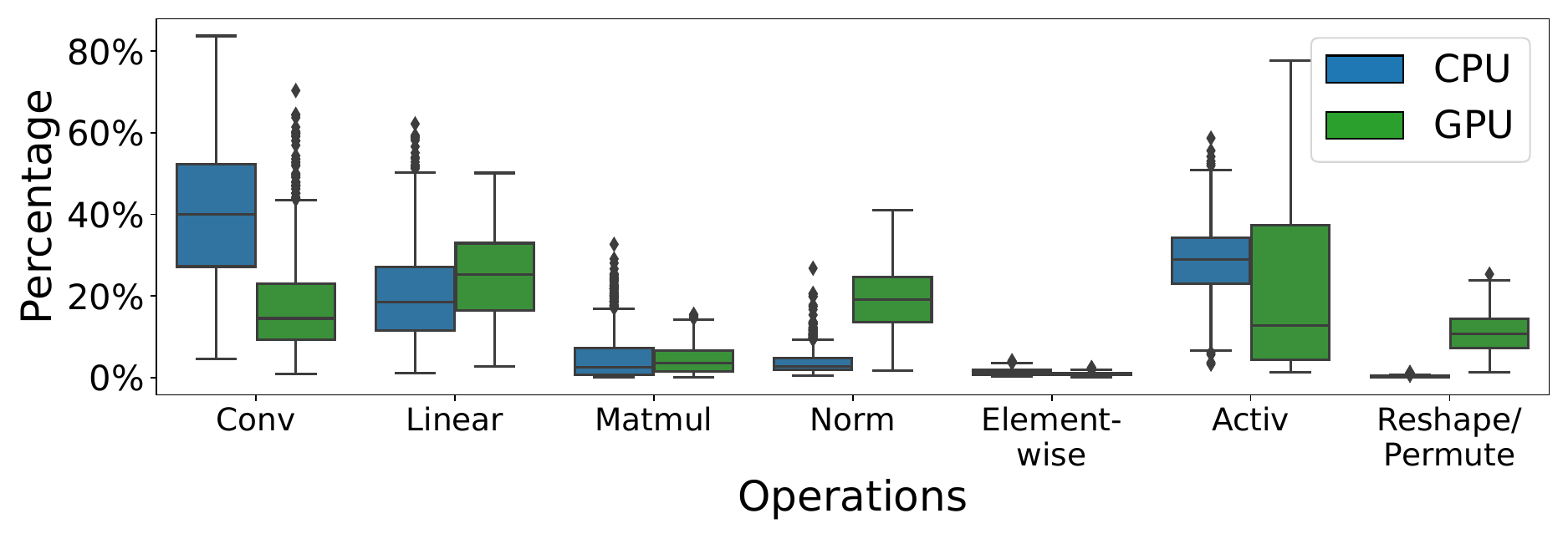}\vspace{-.5em}
	\caption{Latency Breakdown for Synthetic ViTs (PyTorch Mobile)}\label{fig:latency_breakdown_synthetic_gpu_pixel4}
\end{figure}

Based on our search space design, we sample 1000 synthetic ViTs in addition to the 190 real-world ViTs in our dataset, with measurements across 2 ML framework on 6 mobile devices.
Mobile GPUs are widely utilized for inference tasks; however, from our experiments, most real-world ViTs are currently unavailable on mobile GPUs due to limited support of operations in ML frameworks.
Instead, constructing synthetic ViTs facilitates latency measurements on mobile GPUs by substituting unsupported operations in the implementation.
In \cref{sec:result_synthetic,sec:result_realworld}, we further \emph{validate the representativeness of our synthetic ViTs by demonstrating that simple ML predictors trained on these synthetic ViTs can accurately predict the latency of real-world ViTs}.

\cref{fig:latency_breakdown_synthetic_gpu_pixel4} compares the latency breakdown for 1000 synthetic ViTs between CPU and GPU on Pixel 4.
Notably, the convolution operations contribute to a smaller portion of end-to-end latency on GPU, benefiting from the implementations of various efficient Vulkan kernels.
Additionally, the layernorm operations account for a larger portion of end-to-end latency since multiple Vulkan kernels of element-wise operations are dispatched within this operation, incurring more dispatching overhead.
Besides, the latency of reshape and permute operations increases on mobile GPU due to the cost of memory copy from layout transformation~\cite{niu2024smartmem}, while their CPU implementations only update the view of memory format without explicitly memory copy.

\subsection{Prediction Methodology}\label{sec:methodology_predictor}

In order to illustrate the applicability of our dataset~\cite{dataset}, we train latency predictors on the synthetic ViTs and evaluate their accuracy on both synthetic and real-world ViTs.
Similarly to the ML predictors in \cite{li2023predicting} developed for latency prediction of CNNs, we construct latency predictors \emph{for each type of operation}.
Specifically, we focus on common operations in both synthetic and real-world ViTs, including convolution, linear, matmul, element-wise, pooling, normalization, and activation operations.
Notably, we train separate predictors for convolution with different memory formats (i.e., channel-first and channel-last) because of their distinct performance characteristics (as described in \cref{sec:insight_activation}).
Each predictor generates latency prediction for an operation based on the features in \cite{li2023predicting} describing the operation configurations; these features can represent both computational aspects (e.g., FLOPs) and memory access costs (e.g., input and output shapes).
The end-to-end latency is estimated by summing the latency predictions of all operations within a model.
We compare predictions obtained by three ML methods: Lasso, Random Forest (RF) and Gradient Boosted Decision Trees (GBDT), using a loss function of Mean Average Percent Error (MAPE).

\subsection{Prediction on Synthetic ViTs}\label{sec:result_synthetic}

We begin by evaluating our latency prediction on synthetic ViTs, showcasing the applicability of inference estimation for candidate ViT architectures during the NAS process.
Specifically, we divide the dataset of 1000 synthetic ViTs into 900 for training and 100 for testing.

\subsubsection{Comparison of ML Methods}\label{sec:result_synthetic_simple_case}

\begin{table}[t]
\centering
\setlength{\tabcolsep}{0.5em}
\renewcommand\arraystretch{1.1}
\begin{tabular}{c c c c c c c}\toprule
\multirow{2}{*}{\shortstack{\\\textbf{Dataset,}\\\textbf{Framework}}} & \multirow{2}{*}{\shortstack{\\\textbf{Method}}} & \multicolumn{3}{c}{\textbf{Operation MAPE}} & \multirow{2}{*}{\shortstack{\\\textbf{End-to-End}\\\textbf{MAPE}}} \\
\cmidrule(lr){3-5}
& & Conv & Linear & Activ \\
\toprule

\multirow{3}{*}{\shortstack{Synthetic,\\PyTorch}} 
 & Lasso & 17.90\% & 10.69\% & 5.79\% & 15.84\% \tabularnewline
 & RF & 4.31\% & 4.83\% & 4.50\% & 4.10\% \tabularnewline
 & GBDT & 3.93\% & 4.88\% & 4.34\% & 4.44\% \tabularnewline
\midrule

\multirow{3}{*}{\shortstack{Synthetic,\\TFLite}}
 & Lasso & 4.75\% & 4.50\% & 15.97\% & 11.85\% \tabularnewline
 & RF & 1.98\% & 1.97\% & 7.97\% & 4.51\% \tabularnewline
 & GBDT & 2.04\% & 1.81\% & 7.73\% & 4.84\% \tabularnewline
\midrule

\multirow{3}{*}{\shortstack{Real-world,\\PyTorch}}
 & Lasso & 22.06\% & 12.58\% & 11.27\% & 16.78\% \tabularnewline
 & RF & 12.99\% & 11.98\% & 9.77\% & 7.35\% \tabularnewline
 & GBDT & 13.70\% & 10.74\% & 9.77\% & 8.15\% \tabularnewline
\midrule
 
\multirow{3}{*}{\shortstack{Real-world,\\TFLite}}
 & Lasso & 6.19\% & 6.94\% & 12.46\% & 9.31\% \tabularnewline
 & RF & 5.17\% & 6.07\% & 14.66\% & 6.30\% \tabularnewline
 & GBDT & 6.82\% & 5.50\% & 13.26\% & 6.14\% \tabularnewline

\bottomrule
\end{tabular}
\vspace{1mm}
\caption{Mean Average Percentage Errors of CPU Predictions  ($\text{N}_{\text{Train}} = 900$)}
\label{table:prediction_overview_cpu}
\end{table}

\begin{table}[t]
\centering
\setlength{\tabcolsep}{0.5em}
\renewcommand\arraystretch{1.1}
\begin{tabular}{c c c c c c c}\toprule
\multirow{2}{*}{\shortstack{\\\textbf{Dataset,}\\\textbf{Framework}}} & \multirow{2}{*}{\shortstack{\\\textbf{Method}}} & \multicolumn{3}{c}{\textbf{Operation MAPE}} & \multirow{2}{*}{\shortstack{\\\textbf{End-to-End}\\\textbf{MAPE}}} \\
\cmidrule(lr){3-5}
& & Conv & Linear & Activ \\
\toprule

\multirow{3}{*}{\shortstack{Synthetic,\\PyTorch}}
 & Lasso & 10.23\% & 9.33\% & 45.00\% & 27.87\% \tabularnewline
 & RF & 3.30\% & 1.69\% & 1.63\% & 2.32\% \tabularnewline
 & GBDT & 2.06\% & 2.45\% & 7.95\% & 2.12\% \tabularnewline
\midrule

\multirow{3}{*}{\shortstack{Synthetic,\\TFLite}}
 & Lasso & 17.61\% & 9.14\% & 17.19\% & 21.35\% \tabularnewline
 & RF & 8.73\% & 4.93\% & 9.12\% & 8.57\% \tabularnewline
 & GBDT & 9.30\% & 7.12\% & 10.48\% & 8.88\% \tabularnewline

\bottomrule
\end{tabular}
\vspace{1mm}
\caption{Mean Average Percentage Errors of GPU Predictions}
\label{table:prediction_overview_gpu}
\vspace{-2em}
\end{table}

We compare the three ML methods (introduced in \cref{sec:methodology_predictor}) first in a floating-point data representation setting on a large CPU core.
\cref{table:prediction_overview_cpu} summarizes prediction errors (i.e., MAPEs averaged over six mobile platforms in \cref{table:mobile_platforms}) for both end-to-end latency and for three operations significantly contributing to end-to-end latency.
Notably, non-linear ML approaches (RF and GBDT) achieve similarly accurate predictions for both convolution and linear operations (i.e., maximum error of 4.9\%), and thus end-to-end latency (i.e., maximum error of 4.8\%) across PyTorch Mobile and TFLite.
We observe that the errors of convolution operations on PyTorch Mobile are higher than on TFLite, due to the complexity of performance characteristics (as described in \cref{sec:insight_frameworks}).
For this reason, the simple linear method Lasso exhibits significant errors (17.9\% for convolution operations) on PyTorch Mobile due to its limited expressiveness.
In addition, the errors on activation operations are higher, primarily because of mispredictions of GELU operations that behave differently depending on the value of input data (as introduced in \cref{sec:insight_activation}); hence, our ML predictors based on operation configurations (regardless of input values) can hardly capture this behavior and exhibit higher errors.
Nevertheless, since activation operations contribute to a smaller portion of end-to-end latency (compared to convolution and linear operations), our end-to-end latency predictions are still highly accurate.

In addition, we conduct experiments on mobile GPUs to illustrate the high accuracy of our predictions for heterogeneous hardware accelerators.
Due to out-of-GPU-memory issues, it was not possible to collect measurements for our synthetic ViTs using PyTorch Mobile Vulkan backend on Mali G76 and Helio P35. Hence, we only present average prediction errors of two Android GPUs using PyTorch Mobile and four Android GPUs using TFLite in \cref{table:prediction_overview_gpu}.
As with CPU results, all non-linear models (RF and GBDT) obtain comparable accurate end-to-end latency predictions, with maximum error of 2.1\% on PyTorch Mobile and 8.6\% on TFLite; we attribute the higher errors in TFLite to its more diverse implementations of GPU kernel (e.g., using high-performance algorithms to accelerate convolution)~\cite{li2023predicting}.
The linear model Lasso exhibits worse predictions due to the limited capability to capture non-linear relationships between the latency of convolution and the operation features (e.g., kernel size, FLOPs).

Due to lack of space, the rest of this section will focus only on GBDT, because it achieves similarly accurate predictions to RF when there is a larger amount of training data and substantially outperforms RF in the case of limited training data, as discussed in \cref{sec:result_training_size}.

\subsubsection{Feature Importance Analysis}

\begin{figure}[t]
    \centering
    \includegraphics[width=0.52\linewidth]{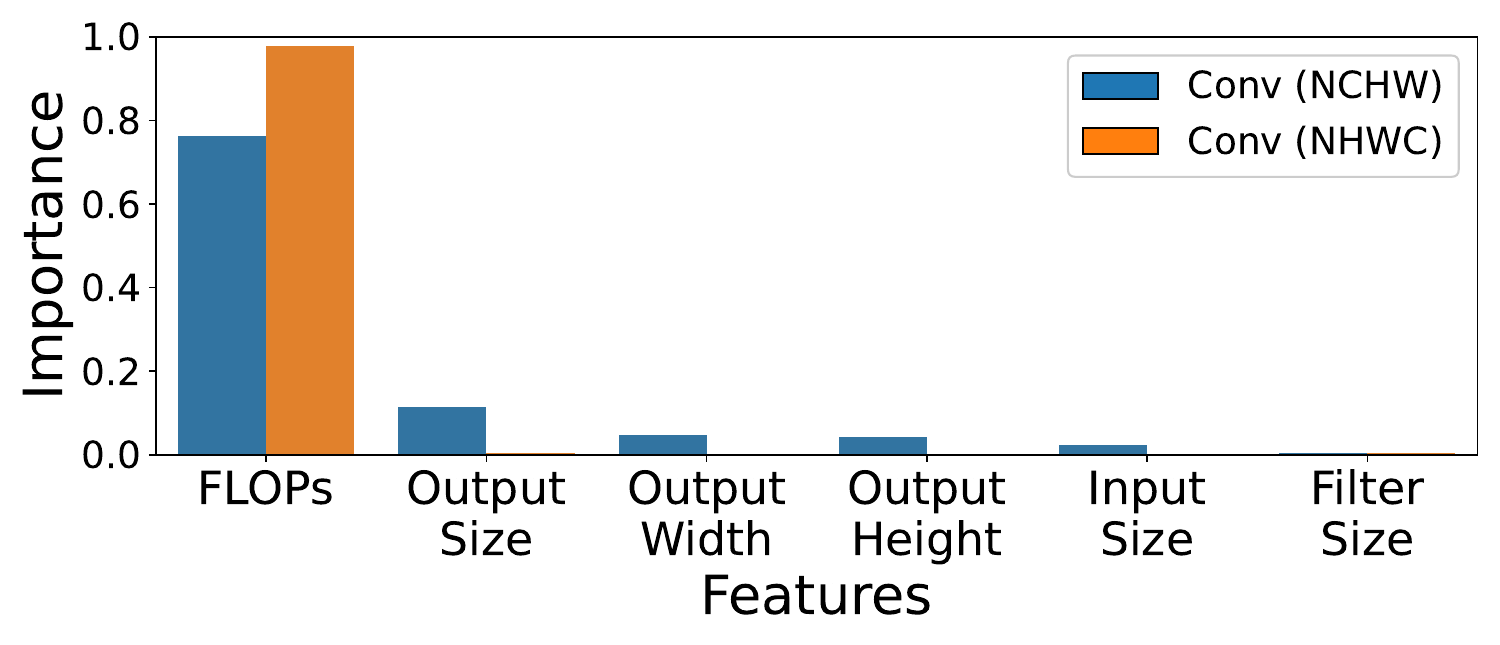}
    \caption{Feature Importance in GBDT Trees}\label{fig:GBDT_feature_importance}
    \vspace{-1.1em}
\end{figure}

Next, we study the importance of various GBDT features to understand critical factors of latency.
Specifically, we use a common metric, Mean Decrease of Impurity (MDI)~\cite{louppe2014understanding}, for feature importance evaluation.
%
%
\cref{fig:GBDT_feature_importance} depicts the features with high importance for convolution operations with different memory formats on Snapdragon 855.
Notably, the latency of convolution operations in channel-last format (NHWC) is essentially determined by FLOPs, while the latency convolution operations in channel-first format (NCHW) is additionally influenced by the sizes of input and output due to the cost of memory copy and reformatting before and after the computation (as detailed in \cref{sec:insight_frameworks}).

\subsubsection{Comprehensive Predictions with GBDT}\label{sec:result_synthetic_complex_cases}

\begin{figure}[t]
    \centering
    \begin{subfigure}[b]{.45\linewidth}
		\centering
		\includegraphics[width=\linewidth]{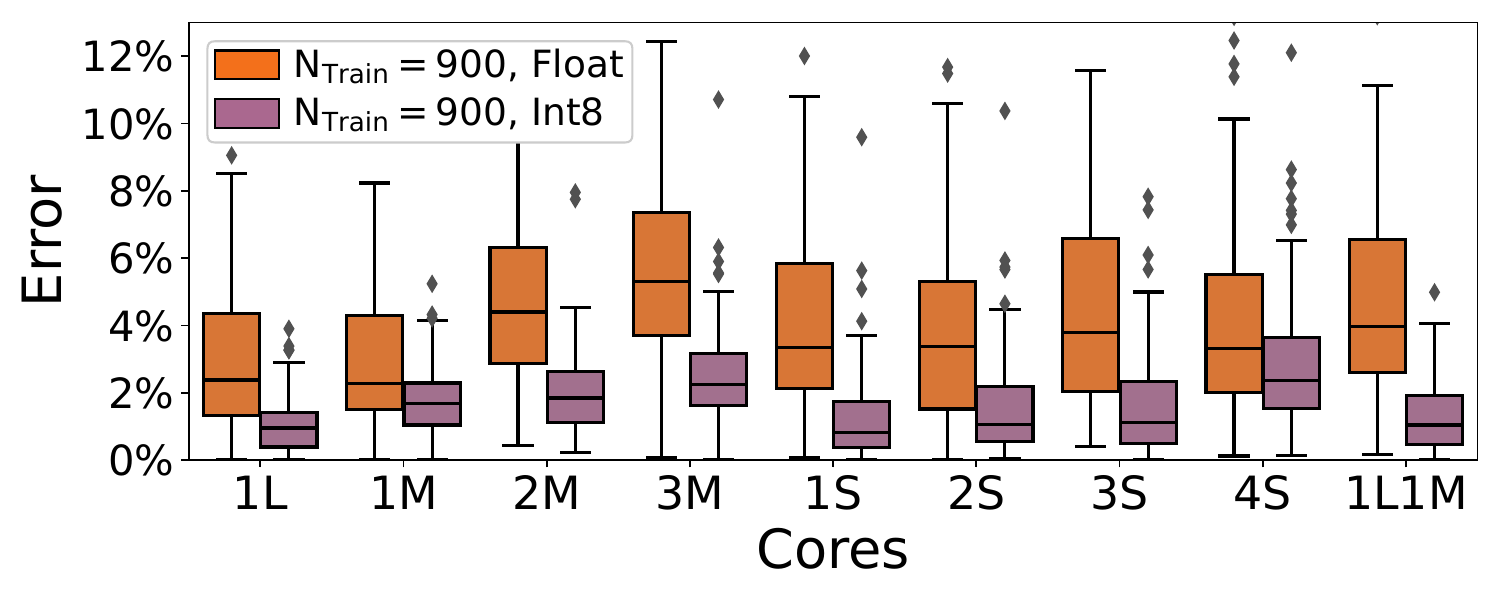}
		\caption{Snapdragon 855 (PyTorch)}\label{fig:predict_torch_synthetic_900_GBDT_weighted_cpu_pixel4_ymax}
	\end{subfigure}
     \hspace{-.8em}
 	\begin{subfigure}[b]{.45\linewidth}
		\centering
		\includegraphics[width=\linewidth]{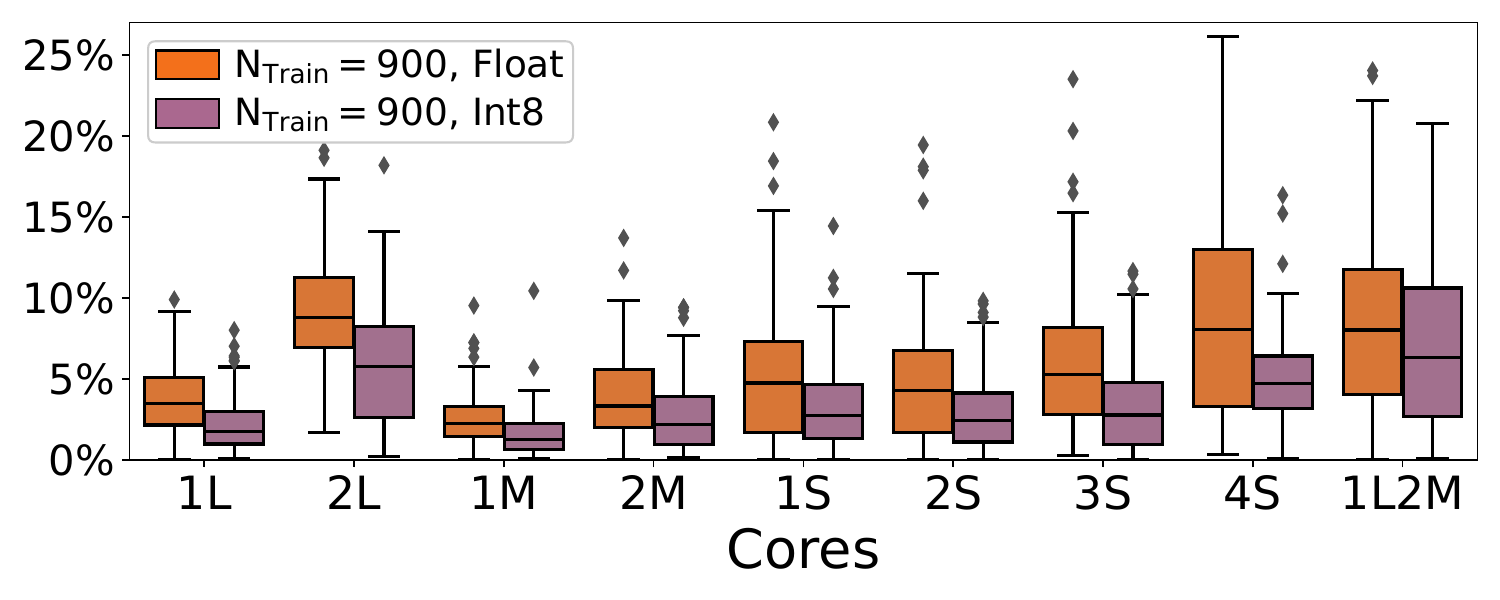}
		\caption{Exynos 9820 (PyTorch)}\label{fig:predict_torch_synthetic_900_GBDT_weighted_cpu_s10_ymax}
	\end{subfigure}
 	\begin{subfigure}[b]{.45\linewidth}
		\centering
		\includegraphics[width=\linewidth]{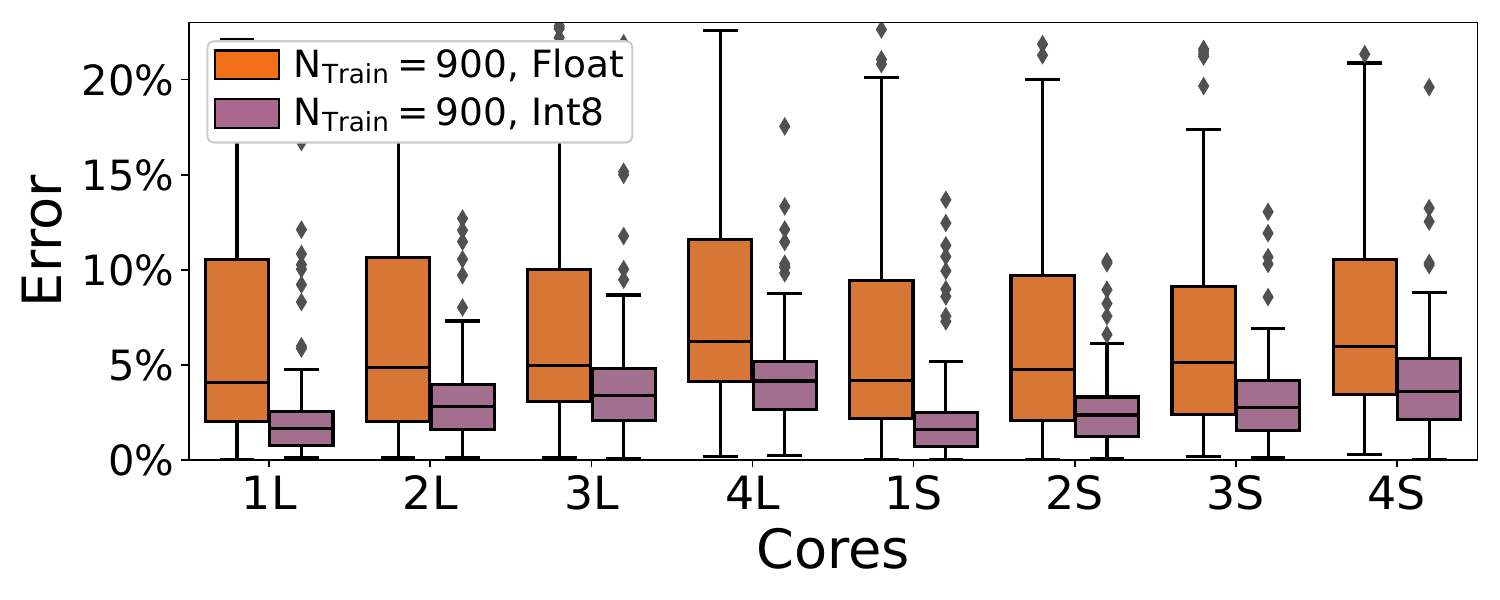}
		\caption{Helio P35 (PyTorch)}\label{fig:predict_torch_synthetic_900_GBDT_weighted_cpu_a03s_ymax}
	\end{subfigure}
     \hspace{-.8em}
 	\begin{subfigure}[b]{.45\linewidth}
		\centering
		\includegraphics[width=\linewidth]{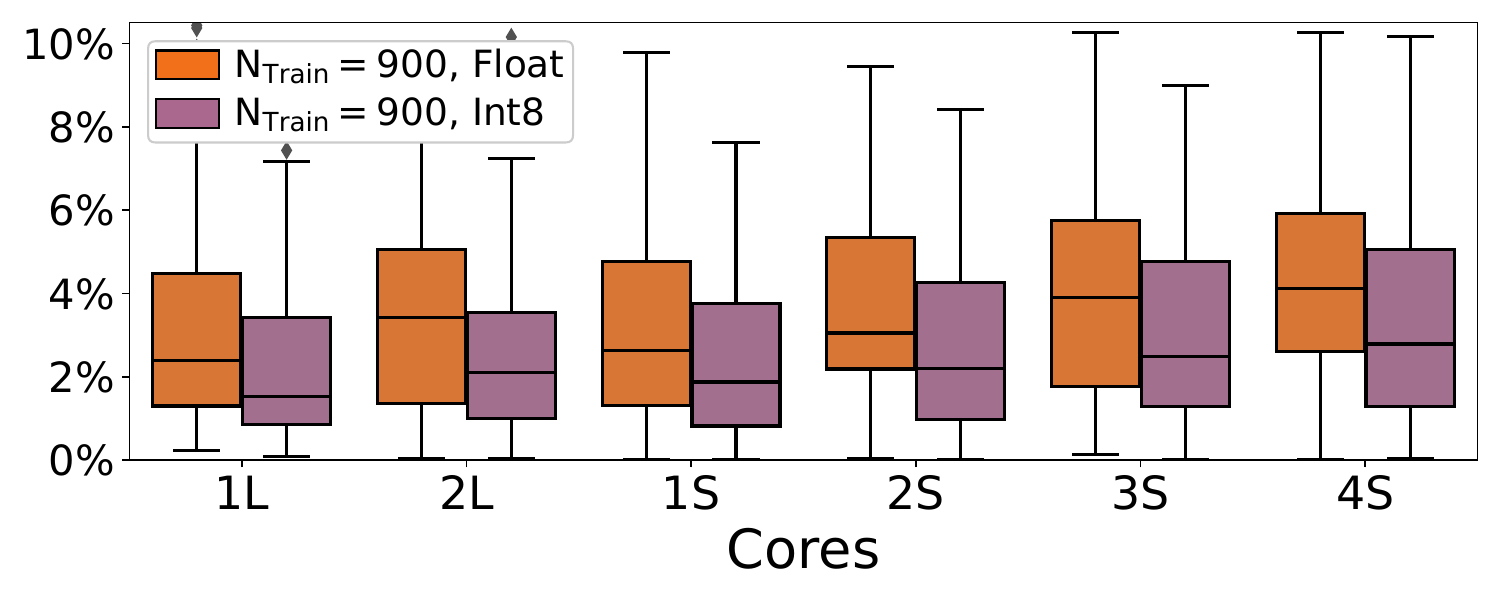}
		\caption{A12 Bionic (PyTorch)}\label{fig:predict_torch_synthetic_900_GBDT_weighted_cpu_iphonexs_ymax}
	\end{subfigure}
	\begin{subfigure}[b]{.45\linewidth}
		\centering
		\includegraphics[width=\linewidth]{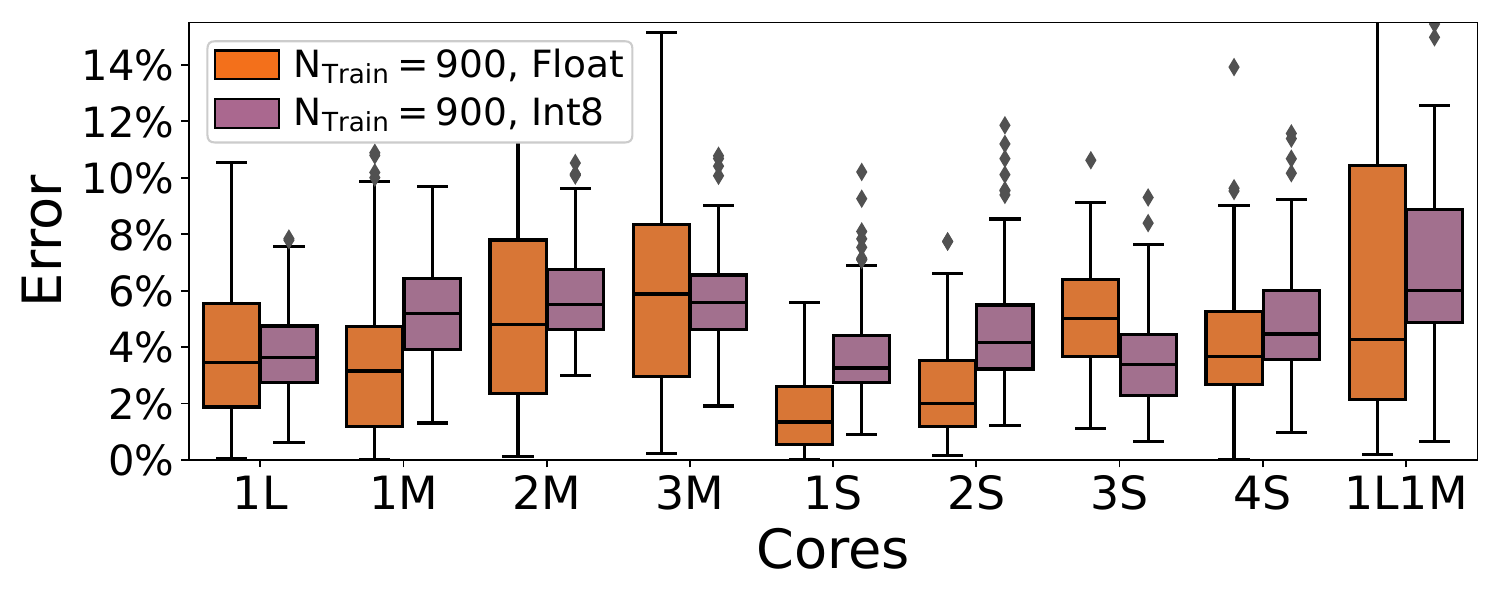}
		\caption{Snapdragon 855 (TFLite)}\label{fig:predict_tf_synthetic_900_GBDT_weighted_cpu_pixel4_ymax}
	\end{subfigure}
     \hspace{-.8em}
 	\begin{subfigure}[b]{.45\linewidth}
		\centering
		\includegraphics[width=\linewidth]{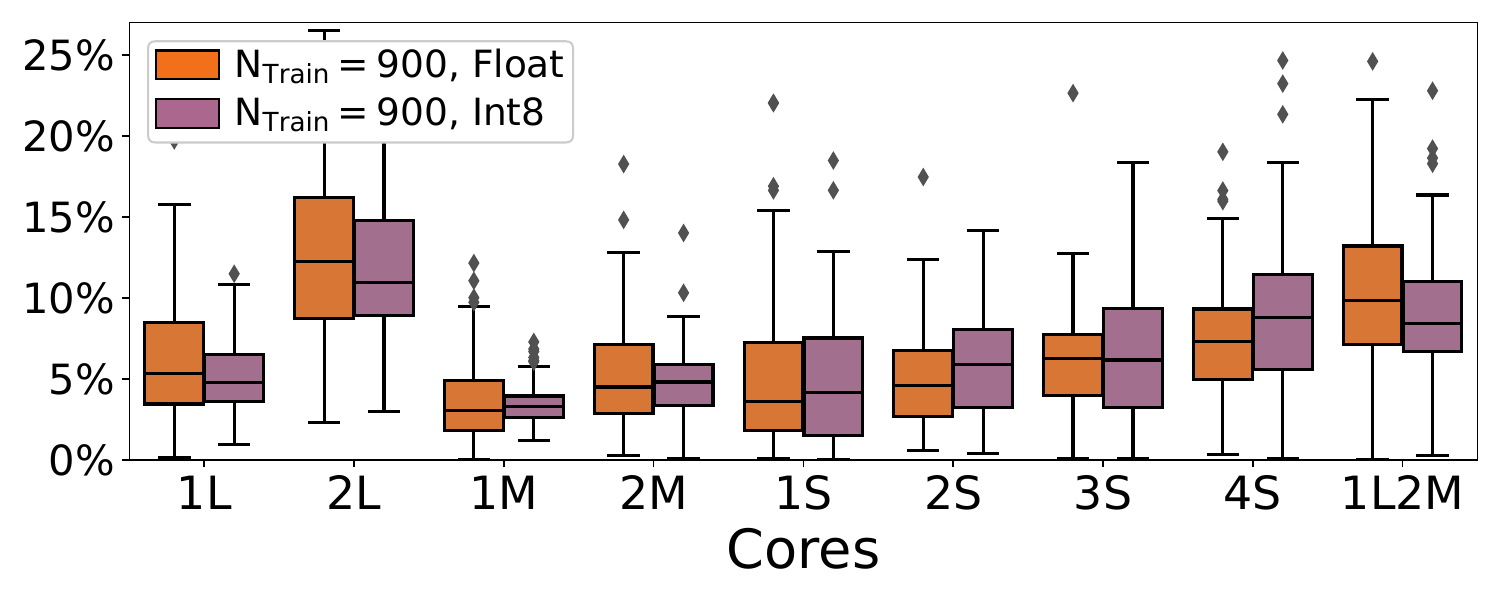}
		\caption{Exynos 9820 (TFLite)}\label{fig:predict_tf_synthetic_900_GBDT_weighted_cpu_a03s_ymax}
	\end{subfigure}
 	\begin{subfigure}[b]{.45\linewidth}
		\centering
		\includegraphics[width=\linewidth]{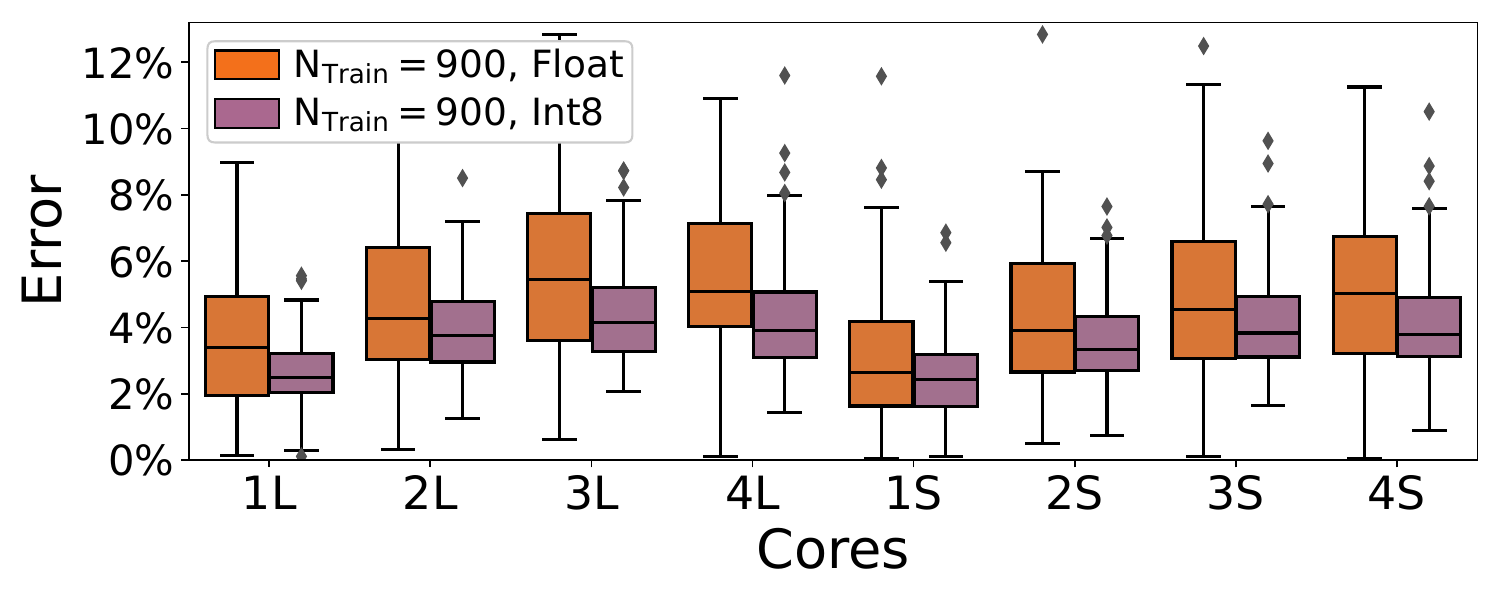}
		\caption{Helio P35 (TFLite)}\label{fig:predict_tf_synthetic_900_GBDT_weighted_cpu_onefusion_ymax}
	\end{subfigure}
     \hspace{-.8em}
 	\begin{subfigure}[b]{.45\linewidth}
		\centering
		\includegraphics[width=\linewidth]{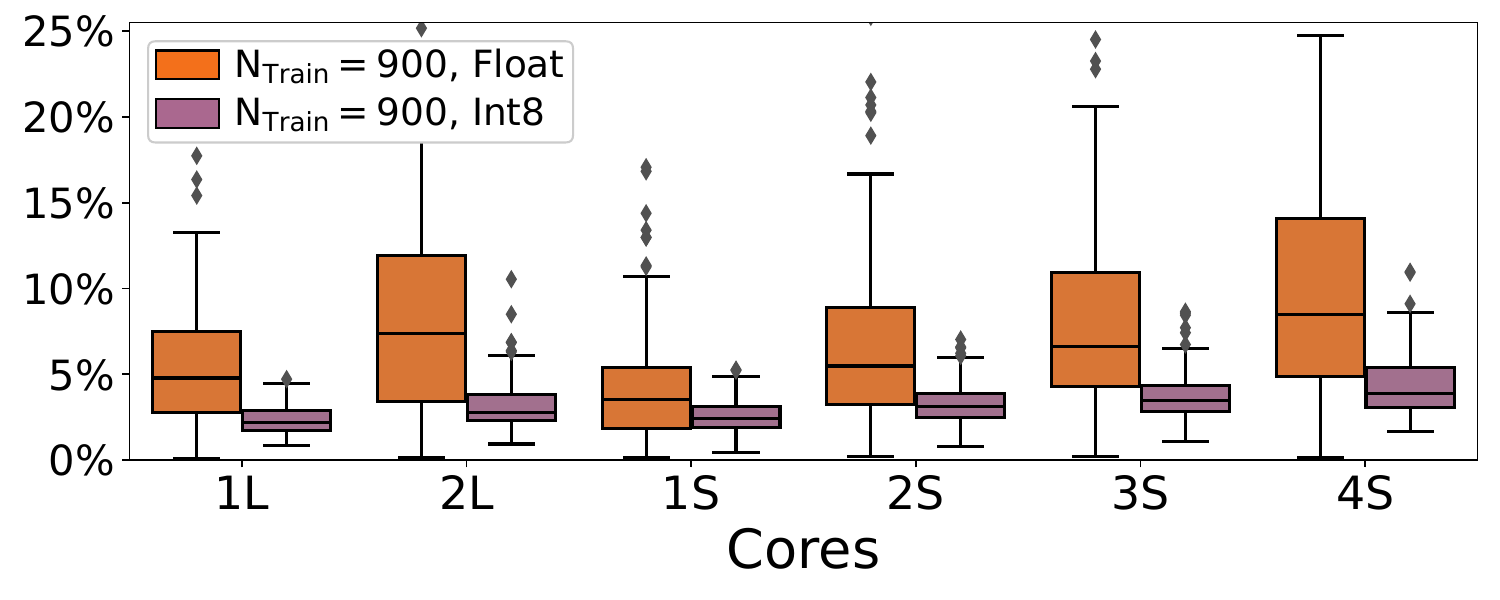}
		\caption{A12 Bionic (TFLite)}\label{fig:predict_tf_synthetic_900_GBDT_weighted_cpu_iphonexs_ymax}
	\end{subfigure}
	\caption{MAPE of GBDT Predictions for Multicore CPUs (Synthetic)}\label{fig:predict_synthetic_cpu}
    \vspace{-.75em}
\end{figure}

Now, we thoroughly investigate prediction results of GBDT across a broad range of practical scenarios, including various data representations and CPU core combinations.
As illustrated in \cref{fig:predict_synthetic_cpu}, GBDT predictions are accurate across all scenarios, with maximum MAPEs of 8.2\% on Snapdragon 855, 12.5\% on Exynos 9820, 6.9\% on Snapdragon 710, 9.1\% on Helio P35, 10.6\% on A12 Bionic, and 11.1\% on A14 Bionic.
Notably, the worst prediction results typically occur when utilizing a large number of cores, especially heterogeneous cores (e.g., 1 large and 2 medium cores on Exynos 9820 in \cref{fig:predict_torch_synthetic_900_GBDT_weighted_cpu_s10_ymax}). This is because using more cores leads to higher resource contention with background jobs on mobile devices, affecting quality of measurements; in particular, multithreading on hybrid cores suffers from overhead due to inter-cluster communication and thread synchronization, resulting in significant prediction errors~\cite{li2023predicting}.
In addition, we observe that the prediction errors are generally lower for integer representations after quantization than floating-point representations, e.g., A12 Bionic in TFLite (\cref{fig:predict_tf_synthetic_900_GBDT_weighted_cpu_iphonexs_ymax}), due to mispredictions on GELUs that account for a larger portion of end-to-end latency in floating-point as compared to integer representations in TFLite, as indicated by the substantial performance improvement for activation functions in \cref{fig:quant_framework_ops}.

\subsection{Predictions on Real-World ViTs}\label{sec:result_realworld}

\begin{figure}[t]
    \centering
    \begin{subfigure}[b]{.45\linewidth}
		\centering
		\includegraphics[width=\linewidth]{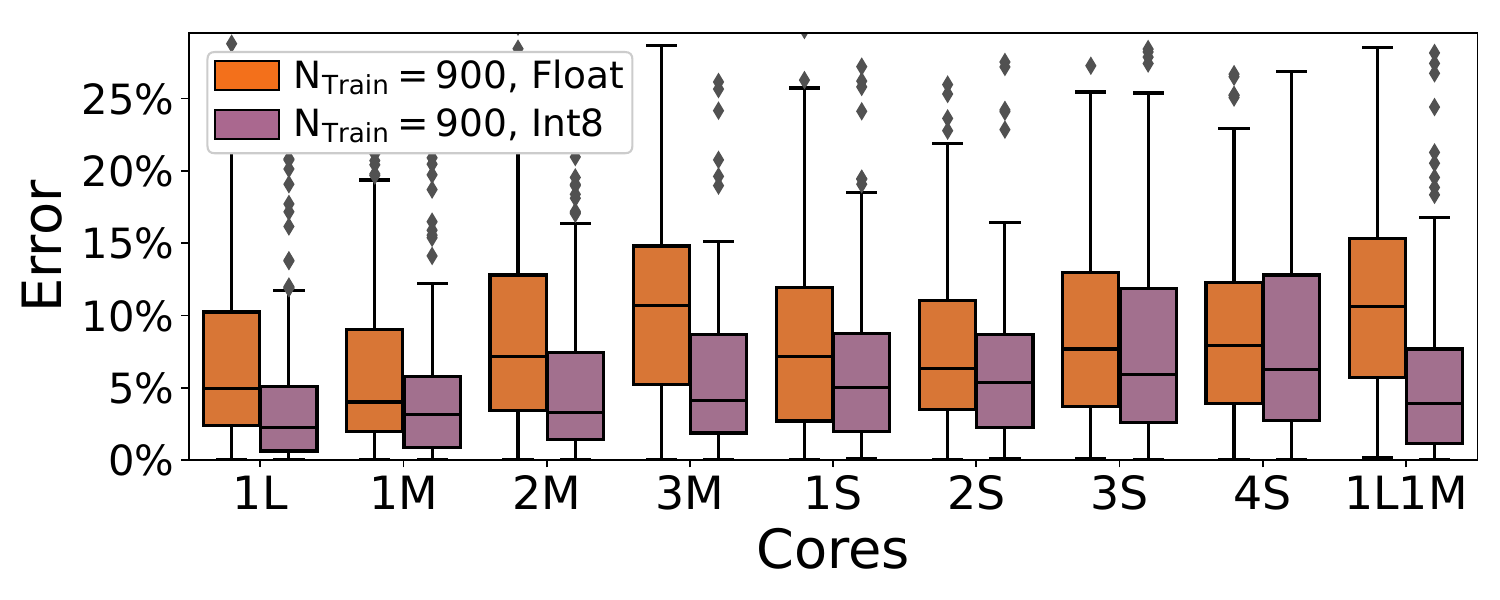}
		\caption{Snapdragon 855 (PyTorch)}\label{fig:predict_torch_realworld_900_GBDT_weighted_cpu_pixel4_ymax}
	\end{subfigure}
     \hspace{-.8em}
 	\begin{subfigure}[b]{.45\linewidth}
		\centering
		\includegraphics[width=\linewidth]{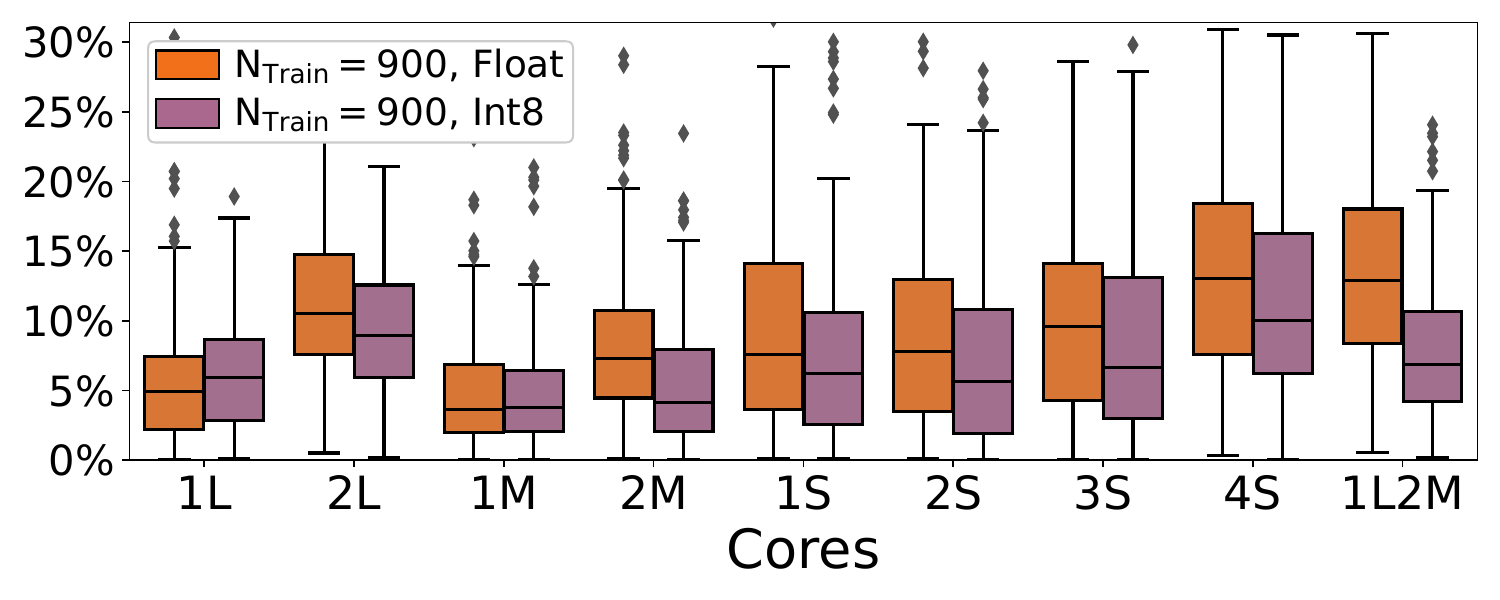}
		\caption{Exynos 9820 (PyTorch)}\label{fig:predict_torch_realworld_900_GBDT_weighted_cpu_s10_ymax}
	\end{subfigure}
 	\begin{subfigure}[b]{.45\linewidth}
		\centering
		\includegraphics[width=\linewidth]{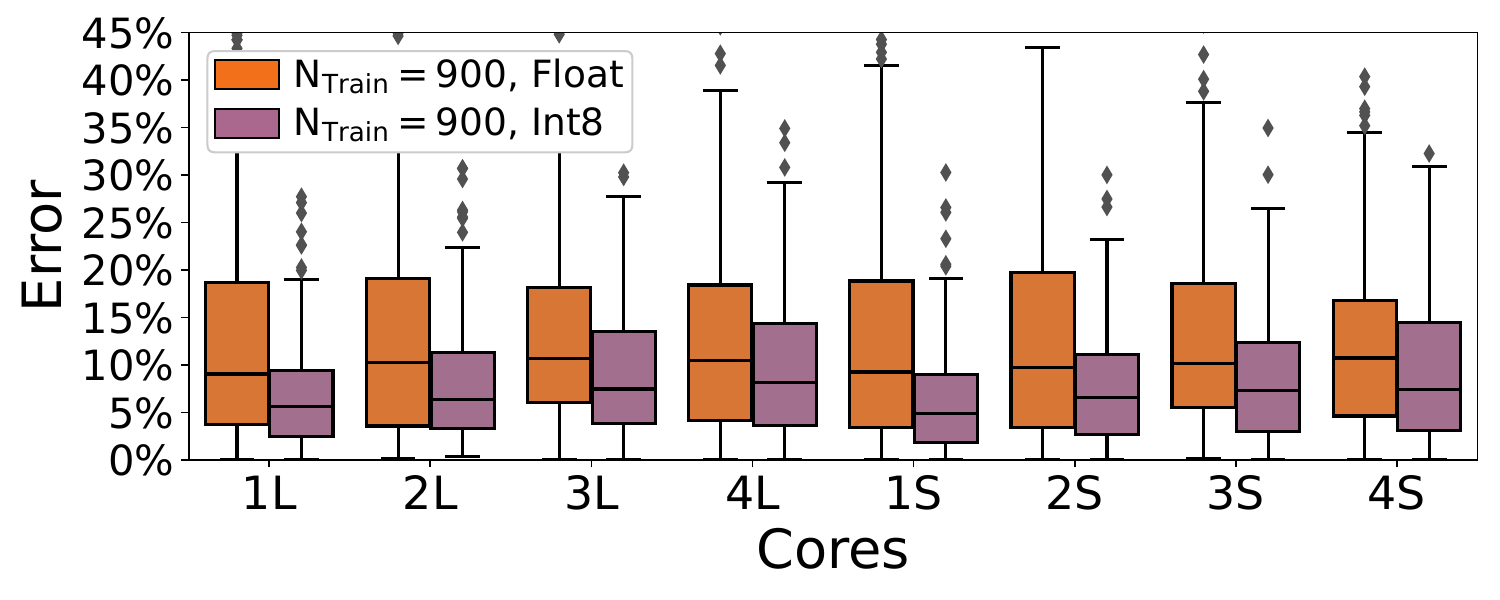}
		\caption{Helio P35 (PyTorch)}\label{fig:predict_torch_realworld_900_GBDT_weighted_cpu_a03s_ymax}
	\end{subfigure}
     \hspace{-.8em}
 	\begin{subfigure}[b]{.45\linewidth}
		\centering
		\includegraphics[width=\linewidth]{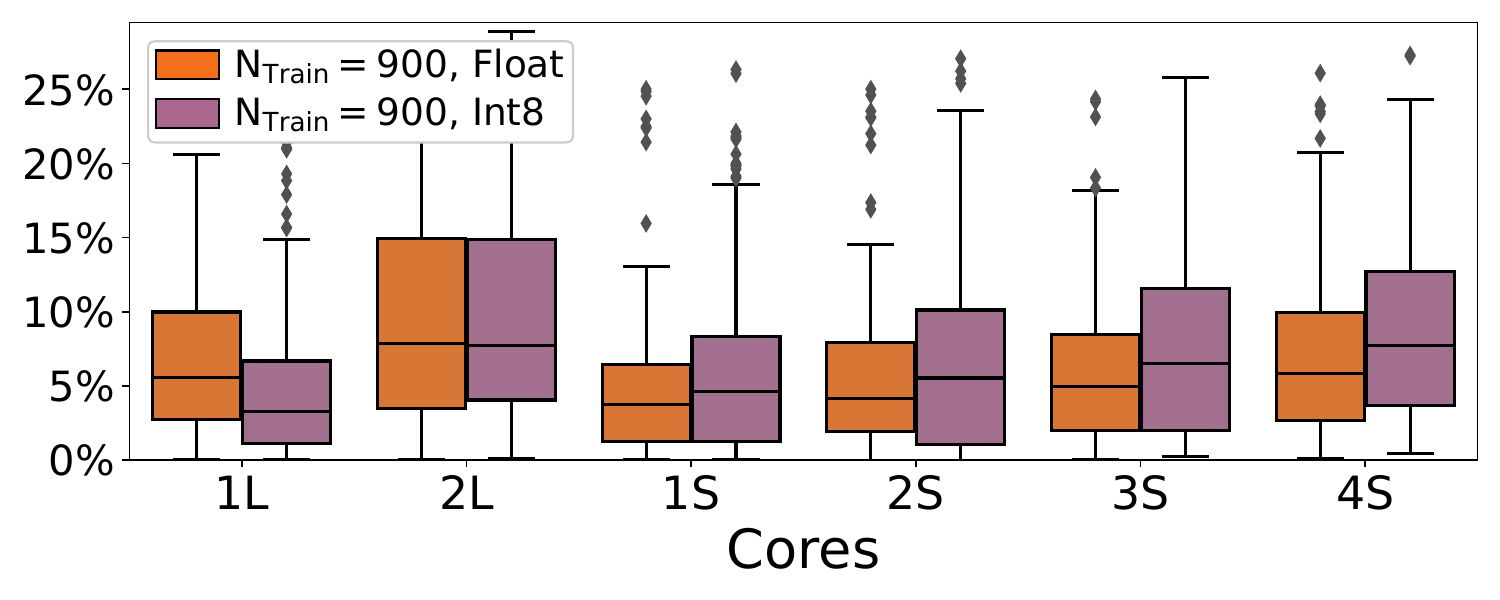}
		\caption{A12 Bionic (PyTorch)}\label{fig:predict_torch_realworld_900_GBDT_weighted_cpu_iphonexs_ymax}
	\end{subfigure}
	\begin{subfigure}[b]{.45\linewidth}
		\centering
		\includegraphics[width=\linewidth]{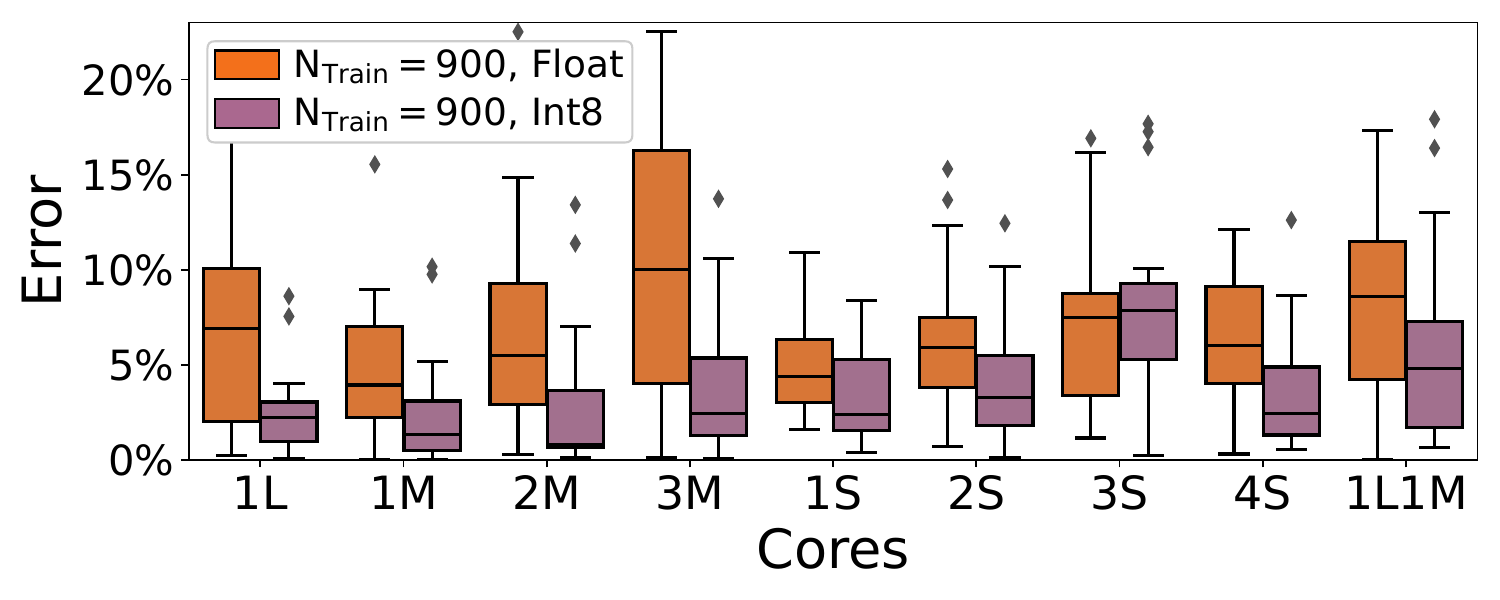}
		\caption{Snapdragon 855 (TFLite)}\label{fig:predict_tf_realworld_900_GBDT_weighted_cpu_pixel4_ymax}
	\end{subfigure}
     \hspace{-.8em}
 	\begin{subfigure}[b]{.45\linewidth}
		\centering
		\includegraphics[width=\linewidth]{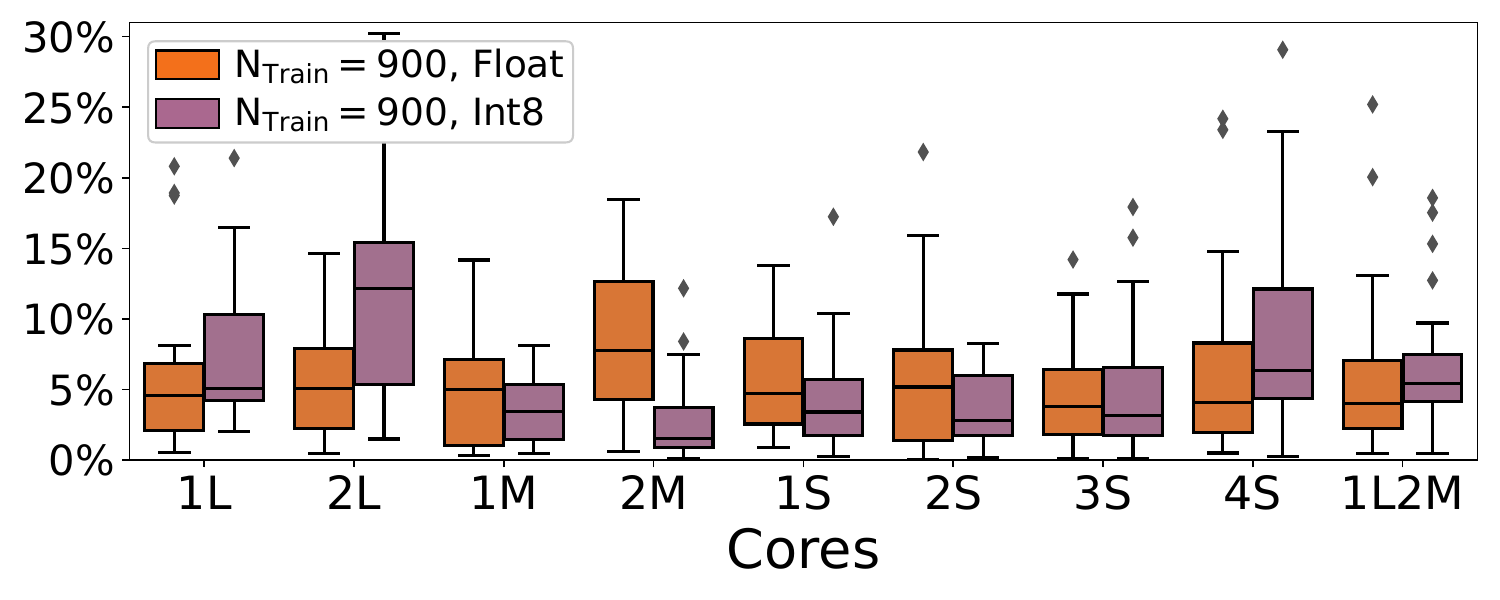}
		\caption{Exynos 9820 (TFLite)}\label{fig:predict_tf_realworld_900_GBDT_weighted_cpu_a03s_ymax}
	\end{subfigure}
 	\begin{subfigure}[b]{.45\linewidth}
		\centering
		\includegraphics[width=\linewidth]{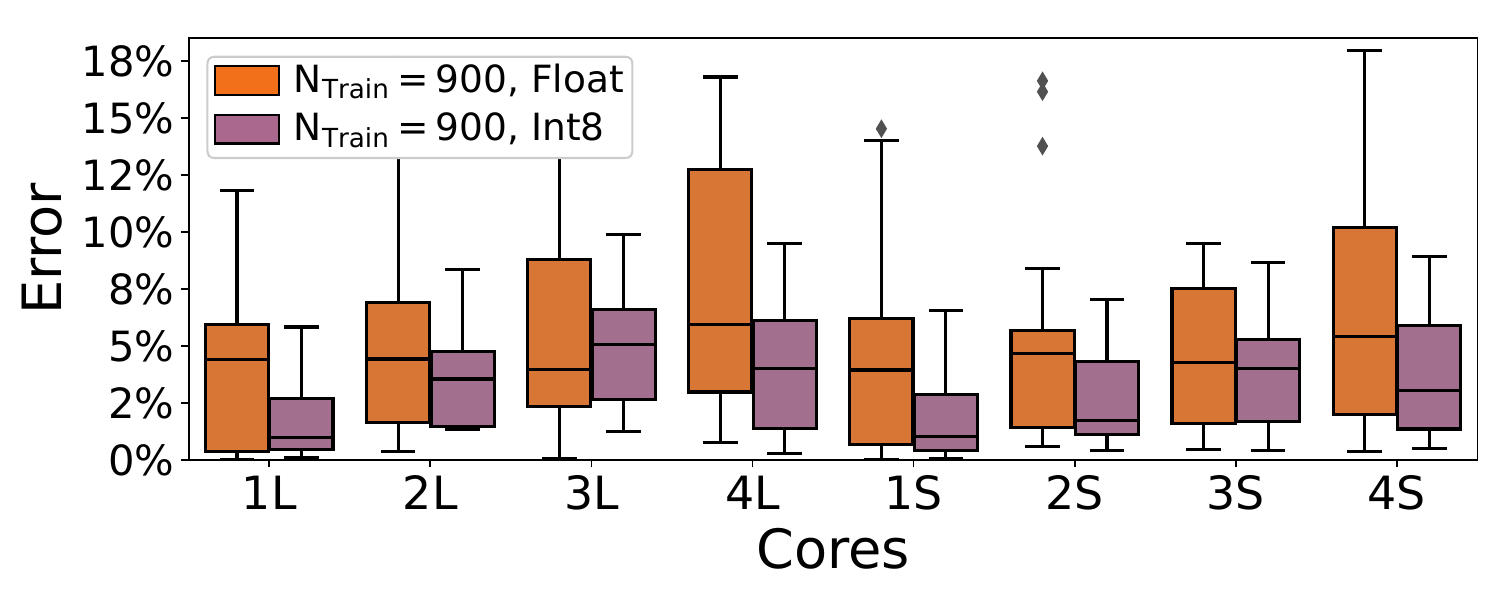}
		\caption{Helio P35 (TFLite)}\label{fig:predict_tf_realworld_900_GBDT_weighted_cpu_onefusion_ymax}
	\end{subfigure}
     \hspace{-.8em}
 	\begin{subfigure}[b]{.45\linewidth}
		\centering
		\includegraphics[width=\linewidth]{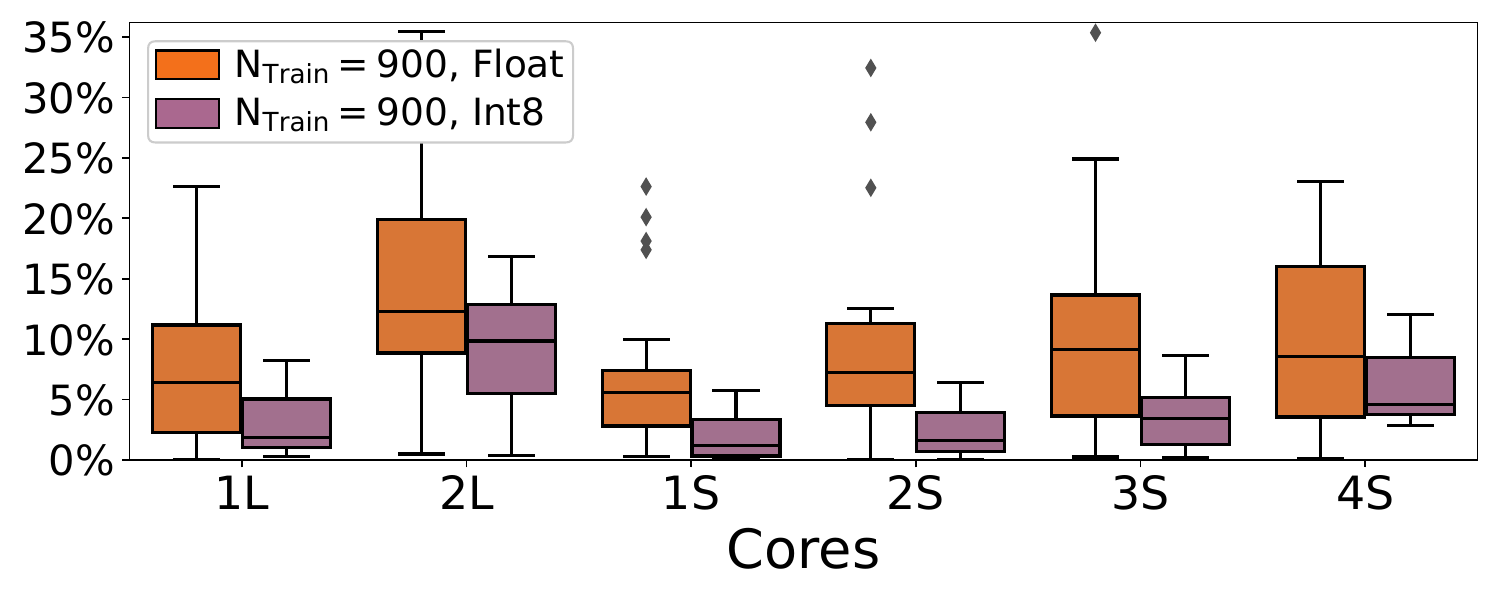}
		\caption{A12 Bionic (TFLite)}\label{fig:predict_tf_realworld_900_GBDT_weighted_cpu_iphonexs_ymax}
	\end{subfigure}
	\caption{MAPE of GBDT Predictions for Multicore CPUs (Real-World)}\label{fig:predict_realworld_cpu}
    \vspace{-1em}
\end{figure}

Next, we conduct experiments applying our ML predictors, trained on synthetic ViTs, to real-world ViTs (described in \cref{sec:result_synthetic}), demonstrating that these predictors can be used to predict latency for SOTA ViT architectures in practical applications such as collaborative inference.

Similarly to the previous setup, we compare the performance of different ML methods on 190 real-world ViTs in PyTorch Mobile and 25 ViTs in TFLite.
As summarized in \cref{table:prediction_overview_cpu}, non-linear methods exhibit higher errors on real-world ViTs than synthetic ViTs, because of the different distribution of operation configurations (i.e., a dataset shift between training and testing).
Specifically, RF and GBDT exhibit accurate end-to-end predictions on real-world ViTs, with errors lower than 8.2\% on PyTorch Mobile and 6.3\% on TFLite.

Additionally, we note that the end-to-end latency prediction error is smaller than that of convolution and linear operations in PyTorch. This is because the MAPE of linear operations is skewed by the mispredictions on certain ViTs that have a few small linear operations (which do not significantly contribute to end-to-end latency); for instance, if we only keep the ViTs in which linear operations account for more than 5\% of end-to-end latency, the MAPE of linear operations is 6.5\% and 6.4\% for RF and GBDT, respectively.
%
We comprehensively evaluate GBDT predictions across diverse scenarios; as shown in \cref{fig:predict_realworld_cpu}, the maximum MAPEs are 11.0\% on Snapdragon 855, 13.3\% on Exynos 9820, 13.7\% on Snapdragon 710, 14.2\% on Helio P35, 13.6\% on A12 Bionic, and 14.9\% on A14 Bionic.
These results illustrate that our pre-trained predictors can be effectively applied to unseen ViT architectures, achieving accurate latency predictions.

\begin{table}[t]
\centering
\setlength{\tabcolsep}{0.5em}
\renewcommand\arraystretch{1.1}
\begin{tabular}{c c c c c c}\toprule
\multirow{2}{*}{\shortstack{\\\textbf{Dataset,}\\\textbf{Framework}}} & \multirow{2}{*}{$\text{N}_{\text{Train}}$} & \multicolumn{3}{c}{\textbf{Method}} \\
\cmidrule(lr){3-5}
& & Lasso & RF & GBDT \\
\toprule

\multirow{4}{*}{\shortstack{Synthetic,\\PyTorch}} & 30 & 16.45\% & 11.89\% & 7.63\% \tabularnewline
\cmidrule{2-5}
 & 100 & 16.02\% & 7.16\% & 4.52\% \tabularnewline
\cmidrule{2-5}
 & 900 & 15.84\% & 4.10\% & 4.44\% \tabularnewline
\midrule
\multirow{4}{*}{\shortstack{Synthetic,\\TFLite}} & 30 & 13.18\% & 9.54\% & 6.26\% \tabularnewline
\cmidrule{2-5}
 & 100 & 12.39\% & 5.88\% & 4.79\% \tabularnewline
\cmidrule{2-5}
 & 900 & 11.85\% & 4.51\% & 4.84\% \tabularnewline

\midrule

\multirow{4}{*}{\shortstack{Real-world,\\PyTorch}} & 30 & 17.52\% & 13.36\% & 10.61\% \tabularnewline
\cmidrule{2-5}
 & 100 & 16.85\% & 9.29\% & 8.94\% \tabularnewline
\cmidrule{2-5}
 & 900 & 16.78\% & 7.35\% & 8.15\% \tabularnewline
\midrule
\multirow{4}{*}{\shortstack{Real-world,\\TFLite}} & 30 & 9.73\% & 11.59\% & 10.43\% \tabularnewline
\cmidrule{2-5}
 & 100 & 9.53\% & 7.04\% & 6.19\% \tabularnewline
\cmidrule{2-5}
 & 900 & 9.31\% & 6.30\% & 6.14\% \tabularnewline

\bottomrule
\end{tabular}
\vspace{1mm}
\caption{MAPE of CPU Predictions using Different Training Dataset Sizes}
\label{table:effect_training_size_cpu}
\vspace{-2em}
\end{table}

\subsection{Effects of Training Dataset Size}\label{sec:result_training_size}

We now study prediction errors as a function of the training dataset size, i.e., the number of ViT architectures in the training data.
As reported in \cref{table:effect_training_size_cpu}, non-linear methods (RF and GBDT) benefit from a larger training data size, while the linear method, Lasso, achieves similar results regardless of training data size due to its simple structure with fewer parameters. 
%
In the case of very limited training data (30 ViTs), GBDT outperforms other methods on synthetic ViTs with low errors (7.6\% on PyTorch Mobile and 6.3\% on TFLite), demonstrating its applicability to NAS. The cost of profiling only 30 ViTs and constructing GBDT predictors is small, compared with comprehensive measurements for thousands of candidate architectures during the search.
%
On the other hand, on real-world ViTs including many unseen operation configurations, Lasso achieves the lowest error (9.7\%) on TFLite when trained on only 30 ViTs, because Lasso only requires small amounts of training data and is more robust to dataset shift; however, for PyTorch Mobile, the non-linearity of performance characteristics (illustrated in \cref{fig:dwconv_performance_torch}) limits the effectiveness of Lasso, resulting in a high error of 17.5\%.

\subsection{Discussion of Limitations}\label{sec:limitations}

We note that inference latency prediction can be hindered by the interference of background tasks running on the mobile platforms. 
In practice, background tasks such as software updates, location services, and network services can consume  computing and memory resources on the mobile platforms unpredictably, leading to the increase of latency of ML inference jobs. These dynamics can affect the accuracy of ML-based latency prediction techniques, since the training data is typically collected under controlled conditions and thus may not account for the existence of background tasks.
We are exploring prediction models for characterizing the inference latency under background tasks as part of future work.
Our initial experiments indicate that this is a challenging problem due to the need to capture the impact of thermal throttling, memory management, and task scheduling across computational units on mobile platforms.

\section{Conclusion}

We provide quantitative analysis on performance characteristics of 190 real-world ViTs on mobile devices through a comparison with 102 real-world CNNs. Some of our main findings include the following: (1) ViTs generally exhibit higher latency than CNNs with similar FLOPs, (2) ViTs exhibit a greater tendency for memory-bound performance compared to CNNs, and (3) the memory requirements of ViTs generally exceed those of CNNs with comparable model sizes.
In addition, we provided insight into factors affecting inference latency of ViTs: (a) leveraging an efficient memory format can significantly improve the computational latency of convolution layers, (b) the values in the input image and model weights affect the latency of GELU activation functions, and (c) various computing libraries within ML frameworks can lead to significant variations in inference latency.

Based on these insights, we developed a dataset~\cite{dataset} of 190 real-world ViTs and 1000 synthetic ViTs with representative building blocks and with measurements from 2 ML frameworks on 6 mobile platforms, across various core combinations and data representations.
From our dataset, we constructed accurate latency predictors for both synthetic and real-world ViTs, demonstrating the applicability to practical problems such as NAS and collaborative inference.
Future work includes exploration of predictions on additional accelerators, such as Neural Processing Units (NPUs) and Digital Signal Processors (DSPs), when more operations of SOTA ViTs are supported by ML frameworks.

%
%
%
\bibliographystyle{splncs04}
\bibliography{main}

\end{document}